\documentclass[lettersize,journal]{IEEEtran}
\usepackage{cite}

\usepackage{graphicx}
\graphicspath{{figures}}
\DeclareGraphicsExtensions{.pdf,.jpeg,.png}

\usepackage{tabularx}
\usepackage{overpic}
\usepackage{siunitx}
\usepackage{acronym}
\usepackage[table,xcdraw]{xcolor}   

\definecolor{green}{rgb}{0.0, 0.5, 0.0}			
\definecolor{gray}{rgb} {0.4, 0.4, 0.4}	
\definecolor{red}{rgb} {0.8, 0.0, 0.0}			
\definecolor{babyblue}{cmyk}{60.0, 40.0, 0.0 ,0.0 }
\definecolor{imesblau}{cmyk}{1.0, 0.78, 0.0, 0.0}
\definecolor{imesorag}{cmyk}{0.0, .6, 1.0, 0.0}
\definecolor{imesgreen}{rgb}{0.78,0.82,0.09}
\definecolor{light_blue}{rgb}{91, 155, 213}

\usepackage{pifont}
\usepackage{cuted}
\usepackage{tikz}
\usetikzlibrary{shapes.geometric, arrows.meta, positioning}

\tikzstyle{startstop} = [rectangle, rounded corners, minimum width=3cm, minimum height=1cm,text centered, draw=black, fill=red!30]
\tikzstyle{process} = [rectangle, minimum width=3cm, minimum height=1cm, text centered, draw=black, fill=blue!20]
\tikzstyle{decision} = [diamond, minimum width=3.5cm, minimum height=1cm, text centered, draw=black, fill=green!20]
\tikzstyle{arrow} = [thick,->,>=stealth]

\usepackage{mathtools}   
\usepackage{amssymb}
\usepackage{siunitx} 
\usepackage[english,algoruled,nofillcomment]{algorithm2e}
	\DontPrintSemicolon
	\SetKwInput{KwIn}{Input}
	\LinesNumbered
	\SetSideCommentRight
	\SetFillComment

	\SetAlFnt{\small}
	\SetAlCapFnt{\small}
	\SetAlCapNameFnt{\small}
	
	\SetCommentSty{mycomfnt}
\usepackage{algpseudocode}
\usepackage{array}
\newcolumntype{?}{!{\vrule width 1pt}}
\usepackage{multirow}
\usepackage{ctable} 
\usepackage{tabularx}
\newcommand{\tablegend}[2][white]{%
\setlength{\fboxsep}{0.2pt}\fcolorbox{black}{#1}{\rule{0pt}{1.5ex}\rule{1.5ex}{0pt}}{\enspace #2}            
}
\usepackage[caption=false,font=footnotesize]{subfig}

\usepackage{stfloats}
\usepackage{verbatimbox}

\usepackage{url}
\usepackage[
	pdftitle={Adaptive Model Predictive Control of a Soft Pneumatic Actuator Using a Physics-informed Neural Network based on Cosserat Rod Theory},
	pdfsubject={},
	pdfauthor={},
	pdfkeywords={},	
        hidelinks
]{hyperref}
\newcommand{\bs}{\boldsymbol}
\newcommand{\ok}{\checkmark}
\newcommand{\no}{$\bs\times$}

\usepackage{enumitem}

\newif\ifhighlightchanges
\ifhighlightchanges
	\newcommand{\hlchanges}[1]{\textcolor{red}{#1}}
\else
	\newcommand{\hlchanges}[1]{#1}
\fi

\newif\ifhighlightchangestwo
\ifhighlightchangestwo
	\newcommand{\hlchangestwo}[1]{\textcolor{red}{#1}}
\else
	\newcommand{\hlchangestwo}[1]{#1}
\fi

\newif\ifhighlightchangesthree
\ifhighlightchangesthree
	\newcommand{\hlchangesthree}[1]{\textcolor{red}{#1}}
\else
	\newcommand{\hlchangesthree}[1]{#1}
\fi

\usepackage{custom_bartholdt}

\newacro{soa}[SOA]{state of the art}
\newacro{dof}[DoF]{degrees of freedom}
\newacro{mae}[MAE]{mean absolute error}
\newacro{rmse}[RMSE]{root mean square error}
\newacro{aed}[AED]{average Euclidean distance}
\newacro{gpu}[GPU]{graphical processing unit}

\newacro{scr}[SCR]{soft continuum robot}
\newacro{cr}[CR]{continuum robot}
\newacro{spa}[SPA]{soft pneumatic actuator}
\newacro{fts}[FTS]{force-torque sensor}
\newacro{ctr}[CTR]{concentric tube robot}
\newacro{smr}[SMR]{soft-material robot}
\newacro{imu}[IMU]{inertial measurement unit}
\newacro{fbg}[FBG]{fiber Bragg grating}

\newacro{cs}[CS]{constant-strain}
\newacro{cc}[CC]{constant-curvature}
\newacro{pcc}[PCC]{piecewise-constant-curvature}
\newacro{pcs}[PCS]{piecewise-constant-strain}
\newacro{gvs}[GVS]{geometric-variable-strain}
\newacro{eom}[EOM]{equation of motion}
\newacro{ode}[ODE]{ordinary differential equation}
\newacro{pde}[PDE]{partial differential equation}
\newacro{bv}[BV]{boundary value}
\newacro{bvp}[BVP]{boundary-value problem}
\newacro{ivp}[IVP]{initial-value problem}
\newacro{cog}[COG]{center of gravity}
\newacro{gff}[GFF]{global-frame formulation}
\newacro{lff}[LFF]{local-frame formulation}
\newacro{ocp}[OCP]{optimal-control problem}
\newacro{nn}[NN]{neural network}
\newacro{rnn}[RNN]{recurrent neural network}
\newacro{lnn}[LNN]{Lagrangian neural network}
\newacro{rl}[RL]{reinforcement learning}
\newacro{pod}[POD]{proper orthogonal decomposition}
\newacro{con}[CON]{coupled oscillator network}

\newacro {fd}[FD]{finite differences}
\newacro{lp}[LP]{lumped-parameter}
\newacro{fem}[FEM]{finite element simulation}
\newacro{iss}[SiS]{single shooting}
\newacro{bdf}[BDF]{backward differentiation formula}
\newacro {cm}[CM]{collocation method}
\newacro{pinn}[PINN]{physics-informed neural network}
\newacro{ddpinn}[DD-PINN]{domain-decoupled physics-informed neural network}
\newacro{pinc}[PINC]{physics-informed neural networks for control}
\newacro{fnn}[FNN]{feed-forward neural network}
\newacro{pso}[PSO]{particle swarm optimization}

\newacro {ssm}[SSM]{state-space model}
\newacro{ut}[UT]{unscented transformation}
\newacro{ukf}[UKF]{unscented Kalman filter}
\newacro{gpr}[GPR]{gaussian process regression}
\newacro{dbn}[DBN]{dynamic Bayesian networks}
\newacro{pinn}[PINN]{physics-informed neural network}

\newacro{mpc}[MPC]{model-predictive control}
\newacro{nempc}[NEMPC]{nonlinear evolutionary model-predictive control}
\newacro{lqr}[LQR]{linear-quadratic regulator}

\newacro{cad}[CAD]{computer-aided design}
\newacro{cas}[CAS]{computer algebra system}

\begin{document}

\title{Adaptive Model-Predictive Control of a Soft Continuum Robot Using a Physics-Informed Neural Network Based on Cosserat Rod Theory}

\author{%
Johann Licher*, Max Bartholdt*, Henrik Krauss, Tim-Lukas Habich, Thomas Seel, Moritz Schappler
\thanks{\texttt{*}Both authors contributed equally to this publication. \\
All authors are with the Institute of Mechatronic Systems, Leibniz University Hannover, 30823 Germany. 
Henrik Krauss is also with the Department of Advanced Interdisciplinary Studies, The University of Tokyo, 153-8904 Tokyo, Japan. 
Johann Licher is also with the Institute of Assembly Technology and Robotics, Leibniz University of Hannover, 30823 Germany. 
Max Bartholdt is also with the Department of Cardiothoracic, Transplantation and Vascular Surgery, Hannover Medical School, 30625 Hannover.
Funded by the Deutsche Forschungsgemeinschaft (DFG, German Research Foundation) under grant no. 405032969.
\texttt{licher@match.uni-hannover.de}}
}



\newif\ifgrammarlyexport

\ifgrammarlyexport
    \tolerance=1
    \emergencystretch=\maxdimen
    \hyphenpenalty=10000
    \hbadness=10000
\else
\fi

\newif\ifcopyright
	 \copyrighttrue

%
%

\ifcopyright
\thispagestyle{empty}
\pagestyle{empty}
{\LARGE IEEE Copyright Notice}
\newline
\fboxrule=0.4pt \fboxsep=3pt

\fbox{\begin{minipage}{1.1\linewidth}  
\textcopyright\,\,2026\,\,IEEE. Personal use of this material is permitted. Permission from IEEE must be obtained for all other uses, in any current or future media, including reprinting/republishing this material for advertising or promotional purposes, creating new collective works, for resale or redistribution to servers or lists, or reuse of any copyrighted component of this work in other works. \\

Accepted to be published in: IEEE Transactions on Robotics (T-RO), 2026.\\

DOI: \href{https://doi.org/10.1109/TRO.2026.3758955}{10.1109/TRO.2026.3758955}
		
\end{minipage}}
\else
\fi

\maketitle

\begin{abstract} 
Dynamic control of soft continuum robots (SCRs) holds great potential for expanding their applications, but remains a challenging problem due to the high computational demands of accurate dynamic models. While data-driven approaches like Koopman-operator-based methods have been proposed, they typically lack adaptability and cannot reconstruct the full robot shape, limiting their applicability.
This work introduces a real-time-capable nonlinear model-predictive control (MPC) framework for SCRs based on a domain-decoupled physics-informed neural network (DD-PINN) with adaptable bending stiffness. The DD-PINN serves as a surrogate for the dynamic Cosserat rod model with a speed-up factor of up to 44,800. It is also used within an unscented Kalman filter for estimating the model states and bending compliance from end-effector position measurements.
We implement a nonlinear evolutionary MPC running at 70 Hz on the GPU. In simulation, it demonstrates accurate tracking of dynamic trajectories and setpoint control with end-effector position errors below 3 mm (2.3\% of the actuator's length). In real-world experiments, the controller achieves similar accuracy. 
\end{abstract}


\section{Introduction}
\IEEEPARstart{S}{oft} continuum robots have emerged as a promising alternative to traditional rigid-link robots in confined, uncertain environments due to their continuous flexure, adaptability, and inherent safety. This characteristic yields unique advantages in applications such as minimally-invasive surgery or inspection~\cite{yang_principles_2018}.
However, the modeling and control of soft continuum robots pose significant challenges because the deformation is not localized at the joints but distributed along the body, resulting in infinite \ac{dof}~\cite{bruder_datadriven_2021}. 


As soft-robotic applications grow in complexity, full-shape awareness and control become essential. To avoid collisions or to deliberately exploit contact with the environment for support, \acsp{scr} must be capable of precisely regulating their entire body configuration~\cite{dellasantina_modelbased_2023}. This capability is particularly valuable in minimally invasive surgery and machine inspection, where whole-body path planning is required~\cite{almanzor_static_2023,habich_intuitive_2023}. Achieving such tasks efficiently further demands controllers that account for the robot’s dynamics. These can produce fast, accurate motion, thereby reducing task durations or enabling efficient biomimetic locomotion in fish-like~\cite{youssef_design_2022} and octopus-like soft robots~\cite{sun_realtime_2025}. Moreover, by exploiting their intrinsic elasticity, soft robots can store and release potential energy, achieving velocities and forces beyond those possible with rigid robots of comparable inertia~\cite{dellasantina_modelbased_2023}. Realizing this potential requires predictive control that incorporates system dynamics.

\begin{figure}[!t]
\centering
        \begin{overpic}[width=0.78\linewidth]{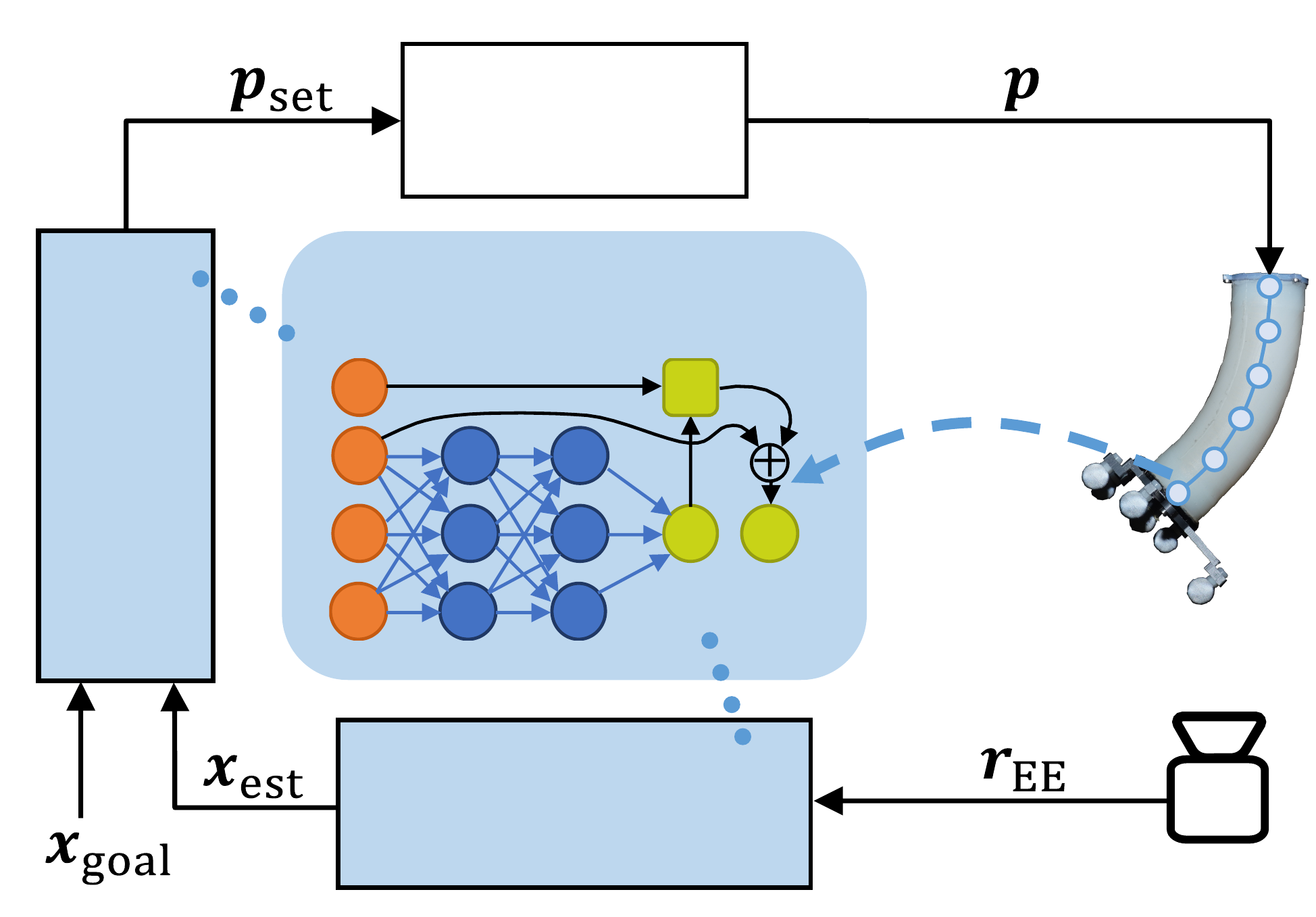}
			\put (34,56.5) {\shortstack[c]{Pressure\\controller}}
            \put (23.5,46) {DD-PINN (\sect{sec_pinn})}
			\put (5,18.6) {\rotatebox{90}{\shortstack[c]{Nonlinear MPC\\ (\sect{sec_NEMPC})}}}
            \put (27.5,5) {\shortstack[c]{State estimation\\(\sect{sec_UKF_design})}}
            \put (66,40) {\shortstack[c]{\textcolor{imesblau}{Cosserat rod} \\ \textcolor{imesblau}{dynamics}\\ \textcolor{imesblau}{(\sect{sec_cosserat_model})}}}
            \put (85,17) {MoCap}
        \end{overpic}
\caption{Overview of the proposed control and estimation framework utilizing a domain-decoupled physics-informed neural network, which is trained on Cosserat rod dynamics.}
\label{fig_graphical_abstract}
\end{figure}
Nonetheless, many existing dynamic models are computationally demanding and unsuitable for real-time control. Physics-informed surrogate models provide a promising solution to this by preserving physical fidelity while enabling efficient computation.

In this work, we present the first demonstration of real-time dynamic control and full-shape state estimation of a \hlchangesthree{single-segment} soft continuum robot using a dynamic Cosserat rod model without prior shape assumptions in real-world experiments (see \figref{fig_graphical_abstract}). 
This is achieved through a \acf{pinn} trained to predict the full Cosserat rod dynamics.

The remainder of this paper is organized into seven sections. 
After an overview of the state of the art and our contributions in \sect{sec_sota} and \sect{sec_contributions}, the Cosserat rod model used in this work is introduced (\sect{sec_cosserat_model}). 
In \sect{sec_pinn}, the physics-informed neural network and its use as a surrogate model are described and experimentally evaluated. \sect{sec_pinn_mpc} presents the nonlinear evolutionary model predictive controller and the unscented Kalman filter, and their experimental evaluation.
Finally, the method is discussed in \sect{sec_discussion}, and conclusions as well as future directions are outlined in \sect{sec_conclusion}.

\section{State of the Art} \label{sec_sota}
The control strategies for soft continuum robots can be divided into \hlchanges{physics-model-based} and data-driven approaches~\cite{dellasantina_modelbased_2023}. 
Recently, an additional class of hybrid control strategies has emerged, combining physics models with machine-learning approaches, for example, in \acp{pinn}~\cite{falotico_learning_2025}.
\hlchangesthree{A secondary criterion for classification is the consideration of the robot's dynamics in control methods}~\cite{grube_comparison_2022}. 

For many applications, the controlled states cannot be measured directly. 
Hence, the state has to be estimated from available measurements to enable closed-loop control. 
In the past, \hlchanges{physics-model-based} and data-driven methods have been used for this task, too. 
While this work is mainly focused on the control of \acp{scr}, we also propose a new model and state estimation method, which are utilized in the control approach. 
This is why we first briefly summarize the state of the art for \ac{scr} modeling (\sect{sec_sota_modelling}) and state estimation (\sect{sec_sota_state_est}) before continuing to physics-model-based, data-driven and hybrid control (\sect{sec_model_based_control_sota} -- \sect{sec_hybrid_approaches_for_control}).
In each of these sections, contributions \wrt the state-of-the-art methods are emphasized and finally summarized in \sect{sec_contributions}.

\subsection{Modeling of Soft Continuum Robots} \label{sec_sota_modelling}

\hlchangestwo{
In the field of soft robotics, a wide variety of modeling strategies have been proposed. 
For the full spectrum of soft continuum robot modeling, the reader is referred to~\cite{armanini_soft_2023}.}

The crux in continuum-robot modeling based on Cosserat rod theory or thereof derived theories, \eg, Kirchhoff-Love beams, is to solve the \acp{pde} in time and space accurately and efficiently.
Models that are suitable for control applications, \eg, state-space models or Lagrangian models, which are \acp{ode}, are desired.
\hlchangestwo{
For this purpose, the one-dimensional spatial domain is discretized with different strategies, which can be classified into energetic and non-energetic approaches \cite{armanini_soft_2023}.
Non-energetic approaches have been investigated to simulate and control continuum robots~\cite{Renda2014, Till2019, Zheng2022, Rucker2022}.
The \textit{single-shooting} method has been a prominent approach due to its comparatively simple implementation, \hlchanges{and a good trade-off between accuracy and computation time~\cite{Till2019}}.
Unlike strain-parametrization-based methods \cite{dellasantina_modelbased_2023, Boyer2021}, no prior assumptions are made on the shape of the robot.
However, it does not achieve the computational efficiency of low-order models, especially if soft-material parameters are considered. 
In fact, it has been reported that the method has inherent drawbacks for \acsp{scr} \hlchangesthree{as the optimization problem is poorly conditioned~\cite{Boyer2023}.
This slows convergence and increases computation time.}}

\textit{In this work}, a \emph{non-energetic approach} based on a simple collocation method is presented.
It tackles the highlighted disadvantages of the single-shooting method and low-order strain parametrization while maintaining state-of-the-art accuracy during experimental validation.
The symbolic implementation using a \ac{cas} and the resulting comparably lightweight expressions allow the derivation of a first-order state-space model formulation in closed form.
Respective output models and Jacobians of the state equation are derived as well to accelerate simulations.

\subsection{State Estimation of Soft Continuum Robots} \label{sec_sota_state_est}
Feedback control requires information on the control variables in real time. 
In some cases, this is inherently difficult to realize with sensors, especially for continuous quantities such as the robot's shape. 
State estimation enables inferring the states from partial observations of the system, \eg, the entire shape of the robot from marker positions along its body.

\hlchangestwo{In recent years, various estimation strategies have been employed for continuum robots.
This includes methods based on kinematics \cite{Martin2022, Sefati2019, Peng2024, Stella2024, Deutschmann2019}, statics \cite{Xu2008, Takano2017, Donat2020, Lilge2022, Zeng2023} and dynamics \cite{DellaSantina2020, thieffry_control_2019, Rucker2022, Ibrahim2021, zheng_estimating_2024, Loo2022, Xavier2022, mehl_adaptive_2024} of the \ac{cr}.
Kinematics-based approaches require a lot of sensors to work accurately \cite{Sofla2021, Cheng2022, Peng2024, Stella2024}. 
Statics-based approaches have mostly been used for classical \ac{cr} manufactured from hard materials and are especially of interest when contact forces at the robot's tip are to be estimated \cite{Xu2008, Takano2017, Donat2021, Alkayas2023}.}
Recent works on Gaussian process regressions with Cosserat rod priors are the first that can be considered hybrid methods that also include model uncertainty \cite{Lilge2022, Lilge2024, Ferguson2024}.
Despite the high accuracy in the presented papers, the estimation methods start to get inaccurate if other than quasi-static motions are performed.
This inherently limits the input band and, therefore, hinders the application to more dynamic tasks, \eg, fast and dynamic manipulation or locomotion.

\hlchangestwo{Many works considering dynamics are evaluated for low-dimensional systems, \eg, the PneuNet finger \cite{Xavier2022, Kim2021,Loo2022} and constrained to or only shown for planar systems \cite{Stewart2022, Tanaka2022, zheng_estimating_2024}. 
Low-order strain approximations have been utilized to design observers \cite{FeliuTalegon2025}, and state estimators \cite{mehl_adaptive_2024}. 
The assumption about the strain modes can be restrictive and is the main drawback of these methods.
Despite the progress in dynamic state estimation, the topic remains challenging.}

\textit{This work} aims to achieve \emph{estimation and control of dynamic states} without the limitation to low-order strain approximations.
To enable \ac{mpc} using a \ac{pinn}, feedback for the state-space model is provided via a nonlinear filter. 
A major advantage of the presented approach is that model parameters that are varied during training can later be adapted online. 
In this way, the bending compliance of the model can be updated with the approach from \cite{mehl_adaptive_2024} to compensate for model inaccuracies.
The fast inference time of the \ac{pinn} allows for computing the unscented Kalman filter for a system with 72 states and one parameter in real-time.


\begin{table}[!tb]
\centering
\addtolength{\tabcolsep}{-0.5em}
\caption{Comparison of \ac{scr} control approaches}
\newcolumntype{N}{>{\centering\arraybackslash}b{0.012\textwidth}}
\newcolumntype{M}{>{\centering\arraybackslash}b{0.0465\textwidth}}
\addvbuffer[0pt 2pt]{\begin{tabular}{@{}cllNMNNN@{}}
Ref.   & Model type                                               & Controller type              & \rotatebox[origin=l]{90}{Dynamics?} & \rotatebox[origin=l]{90}{Real-time?} & \rotatebox[origin=l]{90}{Adaptive?}   & \rotatebox[origin=l]{90}{Real exp.?}  & \rotatebox[origin=l]{90}{Spatial?}  \\ \hline
\rowcolor{imesblau!10} 
\cite{hyatt_model_2020}               & Inextensible PCC                      & Adaptive MPC          & \ok & \ok & \ok & \ok              & \ok      \\
\rowcolor{imesblau!10} 
\cite{katzschmann_dynamic_2019}  & PCC + Augmented rigid body & PD             & \ok & \ok & \no & \ok & \ok \\
\rowcolor{imesblau!10} 
\cite{spinelli_unified_2022}          & PCC + Augmented rigid body     & MPC                   & \ok & 15 Hz    & \no  & \ok              & \ok      \\
\rowcolor{imesblau!10} 
\cite{mei_simultaneous_2023}          & PCC + Lagrangian dyn.      & MPC                   & \ok & \no        & \no  & \no                       & \ok      \\
\rowcolor{imesblau!25} 
\cite{thieffry_control_2019}     & Linearized 3D FEM + POD    & Pole placement & \ok & \ok & \no & \ok & \ok \\
\rowcolor{imesblau!25} 
\cite{soltani_soft_2017}              & Cosserat rod                   & FF MPC                   & \no          & \no     & \no     & \ok                   & \ok      \\
\rowcolor{imesblau!25} 
\cite{shi_position_2024}              & Cosserat rod                   & \hlchanges{FF} MPC                   & \no          & 13 Hz    & \no  & \ok              & \ok      \\
\rowcolor{imesblau!25} 
\cite{caasenbrood_energyshaping_2022} & Cosserat rod                   & Energy-shaping        & \ok & 60 Hz   & \no   & \no                       & \ok      \\
\rowcolor{imesblau!25} 
\cite{danesh_backstepping_2025} & Cosserat rod      & Backstepping   & \ok & \no  & \no & \ok & \no          \\
\rowcolor{imesblau!25} 
\cite{huang_precise_2024}             & Hermite polynomials IK           & Adaptive FF & \no          & 20 Hz   & \ok   & \ok              & \ok      \\
\rowcolor{imesblau!25} 
\cite{renda_dynamics_2024}            & Geometric variable strain      & PD                    & \ok          & \no     & \no    & \no                       & \ok      \\

\rowcolor{imesorag!20} 
\cite{cheney_moldy_2024}              & Feed-forward NN                & NEMPC                   & \ok & 25 Hz   & \no   & \ok              & \ok      \\
\rowcolor{imesorag!20} 
\cite{thuruthel_modelbased_2019}      & Recurrent NN                   & Policy learning       & \ok & 20 Hz   & \no   & \ok              & \ok      \\
\rowcolor{imesorag!20} 
\cite{fang_visionbased_2019}          & Gaussian process               & Visual servoing       & \no          & 20 Hz    & \no  & \ok              & \ok      \\
\rowcolor{imesorag!20} 
        & Linear Koopman operator               & Linear MPC                   & \ok      & 15 Hz & \no & \ok              & \ok      \\
\rowcolor{imesorag!20} 
  \multirow{-2}{*}{\cite{bruder_datadriven_2021}}     & Nonlinear Koopman operator               & FF MPC + LF                & \ok      & \no  & \no & \ok              & \ok      \\
\rowcolor{imesorag!20} 
\cite{haggerty_control_2023}          & Stat. + dyn. Koopman operator               & LQR                   & \ok & 60 Hz   & \no   & \ok              & \ok      \\
\rowcolor{imesorag!40} 
\cite{satheeshbabu_open_2019}         & Model-free                   & Deep-Q RL       & \no        & \no       & \no   & \ok              & \ok      \\
\rowcolor{imesorag!40} 
\cite{morimoto_modelfree_2021}        & Model-free                   & Ensemble RL     & \ok        & \ok & \no & \ok              & \ok      \\
\rowcolor{imesgreen!50} 
\cite{johnson_using_2021}             & PCC-surrogate NN + error-NN    & NEMPC                   & \ok & 100 Hz   & \no  & \ok              & \ok      \\
\rowcolor{imesgreen!50}\cite{grube_open_2024}                & IK-NN + PCC dynamics           & FF          & \ok & \ok & \no & \ok              & \ok      \\
\rowcolor{imesgreen!50} 
\cite{valadas_learning_2025}          & PCS (from video)         & PID-like              & \ok & \no      & \no    & \no                       & \no               \\
\rowcolor{imesgreen!50} 
\cite{stolzle_inputtostate_2024}      & Lagrangian CON (from video)    & PID-like              & \ok & \no       & \no   & \no                       & \no               \\
\rowcolor{imesgreen!50} 
\cite{liu_physicsinformed_2024}       & LNN as PCS surrogate & PD                    & \ok & \no       & \no   & \no                       & \ok      \\
\rowcolor{imesgreen!50} 
Ours           & DD-PINN as Cosserat surrogate                  & Adapt. NEMPC          & \ok & 70 Hz & \ok & \ok              & \ok     \\
\hline
\end{tabular}}
Physics-model-based: \tablegend[imesblau!10]{Geometric} \tablegend[imesblau!25]{Continuum mechanical}\\
Data-driven: \tablegend[imesorag!20]{Regression} \tablegend[imesorag!40]{Reinforcement Learning} $|$~~\tablegend[imesgreen!50]{Hybrid}\\  (FF = feed-forward, IK = inverse kinematics, LF = linear feedback)
\label{tab:control_sota}
\end{table} \vspace{-5pt}

\subsection{Physics-Model-Based Control of Soft Continuum Robots}
\label{sec_model_based_control_sota}
In the context of \hlchanges{physics-model-based} control of \acp{scr}, \textit{discrete}, \textit{geometric}, or \textit{continuum-mechanical} models are typically employed.
\textit{Discrete} methods offer fast prediction and can be used to calculate and consider wrenches along the body. 
However, due to their discrete nature, they are not as accurate as continuum models for elastic structures and require extensive system identification~\cite{alessi_rod_2024}. 

\textit{Geometric} models are characterized by their computational efficiency. 
This is achieved by assuming, for example, constant curvatures and no elongation. 
However, this leads to reduced accuracy under inertial or external forces~\cite{alessi_rod_2024, santina_exact_2019}. 
\hlchanges{Examples of control approaches using the geometric \acf{pcc} model are given in \tabref{tab:control_sota}~\cite{hyatt_model_2020,katzschmann_dynamic_2019,spinelli_unified_2022,mei_simultaneous_2023}.}

\textit{Continuum-mechanical} models encompass \textit{3D mechanical models}\hlchanges{, which represent the detailed 3D structure of the robot and solve it using the finite-element method (FEM), and reduced-order models, such as \textit{rod models}}. 
While the \hlchanges{3D models} are typically the most precise, they also demand the most computational resources. 
Thieffry et al.~\cite{thieffry_control_2019} achieved \hlchanges{real-time control with a 3D model by \ac{pod} and linearization.} However, the linearized model is only accurate around the state used for linearization.

\hlchangestwo{\textit{Rod models} are a structural model-order reduction of 3D continua, which is applicable as long as they are sufficiently slender. 
\hlchanges{\tabref{tab:control_sota} lists multiple control approaches using static~\cite{soltani_soft_2017,shi_position_2024, huang_precise_2024} and dynamic Cosserat rod models~\cite{caasenbrood_energyshaping_2022,danesh_backstepping_2025,renda_dynamics_2024}.}}
The mentioned methods used for control are either limited in terms of accuracy by using simplifying assumptions like (piecewise) constant curvature~\cite{hyatt_model_2020,katzschmann_dynamic_2019,spinelli_unified_2022,mei_simultaneous_2023}, which are inaccurate under inertial forces, or limited in their application by only considering planar motion~\cite{danesh_backstepping_2025}, not considering dynamics~\cite{soltani_soft_2017,shi_position_2024,huang_precise_2024}, not achieving real-time capability~\cite{soltani_soft_2017,renda_dynamics_2024}, or not being validated in real-world scenarios~\cite{caasenbrood_energyshaping_2022}. 
Furthermore, to the best of the authors' knowledge, Cosserat rod models without shape assumptions have not yet been used for closed-loop \ac{mpc} because of the vast computational effort.

\textit{This work} enables this by using a physics-informed surrogate model of a Cosserat rod model without prior shape assumptions, like constant strain. Hence, the method considers dynamics and spatial motion, but still achieves real-time capability, enabling the real-world validation of the \ac{mpc}.

\subsection{Data-Driven Control of Soft Continuum Robots} \label{sec_data_driven_control_sota}

\hlchangestwo{
Data-driven control of soft continuum robots is commonly based on either \textit{supervised regression} or \textit{reinforcement learning}~\cite{laschi_learningbased_2023}.
In supervised regression, the system dynamics are learned from input-output data using models such as feedforward and recurrent neural networks~\cite{cheney_moldy_2024,thuruthel_modelbased_2019}, Gaussian processes~\cite{fang_visionbased_2019}, and Koopman operators~\cite{bruder_datadriven_2021,haggerty_control_2023}. These models can be integrated into advanced control frameworks, including \ac{mpc} and \ac{lqr}.
In reinforcement learning, control policies are learned directly through interaction with the robot or its simulation. 
Existing applications to soft continuum robots include off-policy ensemble learning and Deep-Q learning~\cite{morimoto_modelfree_2021,satheeshbabu_open_2019}, but these approaches have so far been limited to setpoint control.}

\hlchangestwo{
Data-driven methods offer fast inference and require little prior modeling effort. 
However, their performance is generally restricted to operating conditions represented in the training data, often necessitating retraining when system dynamics change~\cite{falotico_learning_2025}.
In addition, collecting sufficient training data can be costly and may accelerate material degradation in physical soft robots~\cite{alessi_rod_2024,morimoto_modelfree_2021,youssef_design_2022}.}

\textit{Our proposed method} mitigates this issue by learning from the dynamic Cosserat rod model rather than from experimental data. Only about \SI{10}{s} of experimental data are required to identify the model parameters. Additionally, the trained \ac{pinn} is adaptive in terms of bending compliance, which increases the method's generalization to altered systems.

\subsection{Hybrid Control Approaches for Soft Continuum Robots}
\label{sec_hybrid_approaches_for_control}
Hybrid approaches integrate \textit{\hlchanges{physics-model-based} and data-driven methods} to leverage the strengths of both. 
Johnson et al.\cite{johnson_using_2021} combine a surrogate \ac{nn} trained on a constant-curvature model with an \ac{nn} trained to estimate the surrogate model’s error. 
Grube et al.~\cite{grube_open_2024} instead use an \ac{nn} to solve the inverse kinematics and combine it with a first-principles dynamic model.
In~\cite{valadas_learning_2025}, a \ac{pcs} representation is learned directly from video data. 
\hlchangestwo{Stölzle and Della Santina~\cite{stolzle_inputtostate_2024} control an \ac{scr} in the latent space, where it is represented by a Lagrangian \ac{con}. 
All of these methods still rely on data for training, so the generalization is dependent on the data amount and quality.}

\textit{Physics-informed surrogate models} offer the dual advantage of \textit{fast prediction} and \textit{improved generalization} beyond the training domain. 
Various methods have been developed that incorporate physics priors into inference, data, architecture, or loss functions of machine learning models applied for control, inverse problems (e.g. identifying a physical system), neural simulations, and more \cite{Hao2022}.
In~\cite{liu_physicsinformed_2024}, \acp{lnn} were employed to control a tendon-driven \ac{scr} in simulation. 
However, 
the underlying \ac{pcc} physics equations limit the overall accuracy.

\hlchangestwo{Bensch et al.~\cite{bensch_physicsinformed_2024} demonstrated that \acp{pinn} can significantly reduce computation time for Cosserat rod statics models compared to first-principles models.}
Ristich et al. \cite{ristich_physicsinformed_2025} use the Cosserat PDEs in physics-informed Koopman operators to obtain a model of \acp{scr} suitable for control. However, neither of them utilized the models for estimation or control.
 
Habich et al.~\cite{habich_generalizable_2025} used a \acf{ddpinn} for nonlinear model predictive control of an articulated soft robot. The \ac{ddpinn} \cite{krauss_domaindecoupled_2024}, also used in this work, utilizes ansatz functions to achieve faster training and more stable predictions than the \ac{pinn}-based control approach presented in~\cite{antonelo_physicsinformed_2024}. 
However, the articulated soft robot controlled in~\cite{habich_generalizable_2025} uses five rigid joints actuated by counteracting inflatable bellows, differentiating it from the continuum nature of \acp{scr}.

\textit{Our method}, however, models continuum robots using a Cosserat rod model without prior shape assumptions and utilizes this model for real-world control and estimation.
Furthermore, the surrogate \ac{pinn} does not include the output model (from states to measured quantities). Hence, in contrast to vision-based methods, like~\cite{valadas_learning_2025,stolzle_inputtostate_2024}, it can be combined with various first-principle output models to utilize different sensors for state feedback without retraining.

\hlchangestwo{\section{Contributions}}
\label{sec_contributions}
The \acf{mpc} approach is often preferred over reinforcement learning because of its good real-world applicability~\cite{hyatt_realtime_2020}. For \acfp{scr}, \ac{mpc} is ``a promising but not widely explored controller"~\cite{alessi_rod_2024}. 
However, so far, only simple models with limiting assumptions have been used for real-time nonlinear \ac{mpc} of soft continuum robots with consideration of the robot's dynamics. 
The main challenge in using more complex models, like the dynamic Cosserat rod model, lies in the high computational cost associated with the models and the multi-step prediction horizon inherent to \ac{mpc}. 
Data-driven control approaches like Koopman-operator methods~\cite{haggerty_control_2023}, on the other hand, usually require more real-world training data and need to be retrained if the system changes.
Armanini et al.~\cite{armanini_soft_2023} highlight neural-network-based surrogate models as a viable means to incorporate complex physical models into real-time control frameworks. 
The existing hybrid approaches, however, rely on rather simple methods with limiting assumptions~\cite{johnson_using_2021,liu_learning_2023}.
This work's main contributions are:
\hlchangestwo{
\begin{enumerate}[label=\textbf{C\arabic*}]
    \item \label{C1} A dynamic Cosserat rod physics model with a collocation-based discretization of the spatial domain that
    has \textit{no prior assumptions on the continuum robot shape}. 
    It enables precise modeling \hlchangesthree{with only \SI{10}{s} of experimental dynamic data and 42 static configurations for parameter identification.}
    \item \label{C2}  An adaptive \ac{ddpinn} as a surrogate for this dynamic Cosserat rod model. This is, to the best of the authors' knowledge, the first time a surrogate model is capable of predicting the entire state of a continuum robot, including strains and velocities, while updating the model's parameters online. Furthermore, it is the largest state-space model (72 states) learned so far using \ac{pinc}. 
    \item \label{C3}  An \acf{ukf} to estimate the model states (internal wrenches and velocities) from the measured end-effector positions using the \ac{ddpinn} as the state-transition model. It also estimates the bending compliance online, allowing the controller to adapt to changes in the real \ac{scr}.
    \item \label{C4} Real-time nonlinear model-predictive \emph{shape} control of a \hlchangesthree{single-segment} soft continuum robot using the \ac{ddpinn} surrogate within the \acf{nempc}.
    This is, to the best of our knowledge, the first time a \ac{pinn} representing the Cosserat rod model with no additional assumptions is used for control of soft continuum robots.
    The proposed controller has been validated in simulation \hlchangesthree{with quantitative evaluation of the full-shape error against ground truth} and in real-world experiments \hlchangesthree{with evaluation of the shape tracking error based on the estimated state}, as well as against a state-of-the-art Koopman controller.
\end{enumerate}}



\section{Cosserat Rod Model --- Local-Frame Formulation} \label{sec_cosserat_model}

\begin{table}[!b]
\centering
\addtolength{\tabcolsep}{-0.5em}
\caption{Symbols and operators}
\addvbuffer[0pt 2pt]{\begin{tabular}{@{}ll|ll@{}}
        Symbol            
        & Description & 
        Symbol            
        & Description     \\ 
        \hline
\rowcolor{imesblau!10}
$\bs{T}(s,t)$ & configuration in $\SE$ & $\loq{\ast}$ & local-frame quantity \\
\rowcolor{imesblau!10}
$s,t$ & material, time & $\bs{\varepsilon}$& relative strain\\
\rowcolor{imesblau!10}
$\position$ & position vector & $\rotation$ & rotation matrix \\
\rowcolor{imesblau!10}
$\loq{\strain}$ & local strain & $\loq{\velocity}$ & local velocity \\
\rowcolor{imesblau!10}
$\strain_0$ & reference strain & $\ell_0$ & reference length \\
\rowcolor{imesorag!40}
$\bs{C}$ & compliance matrix & $\bs{D},\bs{D}_\tau$ & damping matrices \\
\rowcolor{imesorag!40}
$\massmatrix$ & inertia matrix & $\rho,A$ & density, area \\
\rowcolor{imesorag!20}
$\loq{\innerforce}$ & internal force & $\loq{\innermoment}$&internal moment \\
\rowcolor{imesorag!20}
$g$ & gravity constant & $\loq{\externalLoad}$ & external load \\
\rowcolor{imesorag!20}
$\wrenchvector_\mathrm{p}$ & pressure wrench & $p_i$ & chamber pressure \\
\rowcolor{imesorag!20}
$A_{\mathrm{ch},i}$ & chamber area & $\positionchamber$ & chamber offset \\
\rowcolor{imesgreen!50}
$\Expm{\cdot}{\ast}$& exponential map of group $\ast$& $\Adjoint{(\cdot)}$& adjoint map\\
\rowcolor{imesgreen!50}
$\rotation \oplus \bs{\theta} $ &$= \rotation\Expm{\bs{\theta}}{\SO}$ & $\hatmap{(\cdot)}$ & isomorphism of $\SE$\\
\rowcolor{imesgreen!50}
$\sharphat{(\cdot)}$ & isomorphism of $\Adjoint{\SE}$ & $\skw{(\cdot)}$ & isomorphism of $\SO$\\
\hline
\end{tabular}}
\tablegend[imesblau!10]{Kinematics} \tablegend[imesorag!40]{Material} \tablegend[imesorag!20]{Kinetics} \tablegend[imesgreen!50]{Operators}\\
\label{tab:symbols}
\end{table}

This section presents a numerical approach derived from an \ac{ocp}, but not based on single-shooting, to solve the Cosserat rod \ac{pde}.
Instead, the continuity equation is discretized using the implicit midpoint rule. 
This is achieved via a \ac{cas} that allows for symbolic manipulation and differentiation of the obtained equations.
Consequently, the \ac{ocp} becomes a set of nonlinear equations representing the boundary values and continuity constraints.
A fully implicit state-space model is thereby obtained, which will later serve as a loss function for the training of the \ac{ddpinn}.
However, for simulation and analysis, it is more convenient to transform the system into an explicit first-order model of the form $\diff{\state}{t} = \bs{f}(\state, \bs{u}, \parameterCosserat)$ with states $\state$, input $\bs{u}$ and partially unknown parameters $\parameterCosserat$ of the \underline{c}ontinuum \underline{r}obot.
Moreover, a state-space model applied in estimation and control requires a function $\bs{h}$ that describes its outputs $\bs{y}$.
Both the state and output equations are derived using symbolic computation.
This allows for the application of any standard solver, such as Matlab's \texttt{ode15s}, which is suitable for stiff \acp{ivp}.
In contrast to the fixed time step used in the \ac{bdf} method for single shooting, these solvers employ variable time steps, thereby achieving higher accuracy. 
This is especially important if high input dynamics are considered. 


Sec.~\ref{sec_cosserat_rod_theory} reviews Cosserat rod theory, including actuation (Sec.~\ref{sec_cosserat_actuation}) and gravitational loads (Sec.~\ref{sec_cosserat_gravitation}).
Sec.~\ref{sec_collocation_method} details the numerical method, and Secs.~\ref{sec_state_space} and~\ref{sec_output_models} reformulate the discretized model into explicit state and output equations.
Finally, Sec.~\ref{sec_linerarization} derives the Jacobian of the state equations and briefly analyzes the linearized model.
Table~\ref{tab:symbols} summarizes the symbols and operators.

\subsection{Cosserat Rod Theory}
\label{sec_cosserat_rod_theory}
The configuration of the robot's deformed body at time $t$ is described by a field of homogeneous transformation matrices $\htf\st \in \SE$  containing positions $\position\st$ and orientations $\rotation\st\in\SO$ of the rod's cross-section, 
\hlchanges{
where $s$ denotes the curvilinear spatial coordinate along the rod.}
For the sake of brevity, the dependencies of $s$ and $t$ are omitted in the following.
The derivatives of the configuration \wrt the spatial and temporal domain are defined by the local tangential spaces of $\SE$
\begin{align}
	\label{eq.modelling.rodkinematics}
	\diff{\htf}{s} &= \htf \hatmap{\loq{\strain}}, &\diff{\htf}{t} = \htf \hatmap{\loq{\velocity}},
\end{align}
with local strains $\loq{\strain} \in \VectorSpace{6}$,  local velocities $\loq{\velocity}\in \VectorSpace{6}$ and the isomorphism $\hatmap{(.)}:\VectorSpace{6} \rightarrow \frak{se}(3)$. 
The definition in \eqref{eq.modelling.rodkinematics} holds for the reference configuration ${\htf}_0$ with $\strain = \strain_0$, which describes the shape of the rod as the derivative of ${\htf}_0$ with respect to $s$.
In this work, without loss of generality, it is assumed that $\strain_0 = \trans{\inlineMatrix{0&0&1&0&0&0}}$, which describes an initially straight rod. 

As the numeric treatment of the \ac{pde} differs from other works, \eg, energy-based methods in \cite{Boyer2021}, the notation slightly varies and is based on the intrinsic formulation used in \cite{Artola2019}.
The rod is described by a field $\bs{x}(s,t): [0, \ell_0] \times \mathbb{R}_{+} \rightarrow \mathbb{R}^{12}$ with the rod's reference length $\ell_0$. 
The first six components of the state vector represent the linear and angular velocities in local coordinates, given by $\bs{x}_1 = \loq{\velocity} = \trans{[\trans{\loq{\linvel}}, \trans{\loq{\angvel}}]}$. 
The remaining six states,
\begin{equation}
\label{eq.state_x2}
    \bs{x}_2 = \bs{C}^{-1} \bs{\varepsilon}\text{,}
\end{equation}
correspond to the internal forces and moments due to elastic coupling of the material. 
Here, $\bs{C}$ is the compliance matrix, and \hlchanges{$\bs{\varepsilon} = \loq{\bs{\xi}} - \loq{\bs{\xi}}_0$} represents the relative strain. 


When $\bs{x}_2$ is inserted into the Kelvin–Voigt material model, this leads to the constitutive equation
\begin{equation} 
\label{eq_material_law}
\wrench =\boldsymbol{x}_2+\boldsymbol{D}_\tau \diff{}{t}\boldsymbol{x}_2 \quad \text {with} \quad \boldsymbol{D}_\tau=\boldsymbol{D} \boldsymbol{C},
\end{equation}
where $\loq{\bs{n}}$ denotes the local forces, $\loq{\bs{m}}$ the local moments, $\bs{D}_\tau$ is a damping matrix containing eigenvalues of the material's transfer function \cite{Artola2019}, and $\bs{D}$ is the standard damping matrix used in Kelvin-Voigt materials for Cosserat rods. 
Using this in the balance equation of an infinitesimal rod section leads to 
%
    \begin{multline}
	 \diff{\left( \stateB + \dtau \diff{\stateB}{t}\right)}{s}
		-\trans{\sharphat{
				\underbrace{
					\left( 
				\bs{C}\stateB + \loq{\strain}_0
				\right)
				}
			_{=\loq{\strain}}
			} } \left(\stateB + \dtau \diff{\stateB}{t}\right)\\[-3pt]
	=
	\massmatrix \diff{\stateA}{t} 
	- \trans{\sharphat{\stateA}}\massmatrix\stateA
	-\loq{\externalLoad}
    \label{eq.modelling.lff_balance_equations}
    \end{multline}
%
with the distributed external wrench $\loq{\bs{\zeta}}_\ind{ext}$ and the cross-sectional inertia tensor $\bs{M}$. 
The ${\sharphat{}}$ operator is the isomorphism associated with the Lie group of the adjoint $\Adjoint{\SE}$.
The remaining six PDEs are given by the compatibility equation
\begin{align}
\begin{split}
	\diff{\stateA}{s}
	= 
	\bs{C}
	\diff{\stateB}{t}
		-\sharphat{\loq{\strain}_{0}} \bs{x}_1 
	+ \sharphat{\stateA}
	\bs{C}
	\stateB
\end{split}.
	\label{eq.modelling.lff_compatibility}
\end{align}
Additionally, the required boundary values for a cantilevered beam are given by
\begin{align}
	\stateA(0,\cdot) = \zero,\, \stateB(\ell_0,\cdot) + \dtau \diff{\stateB(\ell_0,\cdot)}{t} = \zero, \,	\htf(0,\cdot) = \htf[base]
	\label{eq.modelling.cosserat.boundaryvalues}
\end{align}
with the configuration at the rod's base $\htf[base]$.

\subsection{Actuation}
\label{sec_cosserat_actuation}
The continuum robot is driven by air pressure. 
Three air chambers, which are symmetrically aligned around the lateral axis of the rod, are connected to a source of pressure via pneumatic lines and valves. 
\hlchanges{
It follows that the robot is underactuated because it has more \acp{dof} than inputs. 
This is addressed by the \ac{mpc}, which---in simple terms---finds optimal inputs to a state trajectory.
}
The air pressure is always acting normal to the cross-section, which results in a locally defined wrench $\loq{\wrenchvector}_\mathrm{p}$.
This introduces an unsteadiness in the internal forces and moments at the beginning and the end of the air chambers \cite{Till2019}.

According to \cite{Till2019}, the local force and moment applied by air pressure at the ends of the $i$-th chamber are 
\begin{equation}
	\loq{\wrenchvector}_\mathrm{p} = \sum_{i = 1}^{3}p_i A_{\mathrm{ch},i} \Pmatrix{\bs{e}_z\\ \skw{\positionchamber} \bs{e}_z }
\end{equation}
with pressure $p_i$ and local position vector $\positionchamber$ of the $i$-th air chamber, pointing from the cross-section's origin to the geometric centre of each chamber, with cross-sectional area $A_{\mathrm{ch},i}$ and its normal unit vector $\bs{e}_z$. 
The operator $\skw{\ }:\VectorSpace{3}\mapsto\VectorSpace{3\times3}$ maps its argument to a skew-symmetric matrix so that $\skw{\bs{a}}\bs{b} = \bs{a}\times\bs{b}$ and is at the same time the isomorphism of the $\SO$ group.

The {change in force and moment by actuation with respect to the local frame} is then given by the derivative 
%
\begin{equation}
	\diff{\loq{\wrenchvector}_\mathrm{p}}{s} = \loq{\externalLoad[,p]} = \trans{\sharphat{\loq{\strain}}}
	\loq{\wrenchvector}_\mathrm{p}\, .
\end{equation}
The air chambers within the actuator have a starting point $s = s_\mathrm{p,s}$  and end point $s = s_\mathrm{p,e}$.
The local wrench, due to pressure, is applied at these points, which is expressed as
\begin{align}
	\wrench(s_\mathrm{p,s/e},\cdot) = \lim_{\tilde{s}\rightarrow 0} \wrench(s_\mathrm{p,s/e} + \tilde{s},\cdot) \mp \wrenchvector_\mathrm{p}\, .
	\label{eq.modelling.cosserat.transitioncondition}
\end{align}
This defines additional constraints at the distal ends of each actuator that have to be handled in the numerics later on.

%

\subsection{Gravitational Loads}
\label{sec_cosserat_gravitation}
\begin{figure}
	\centering
	\includegraphics[width=0.9\linewidth]{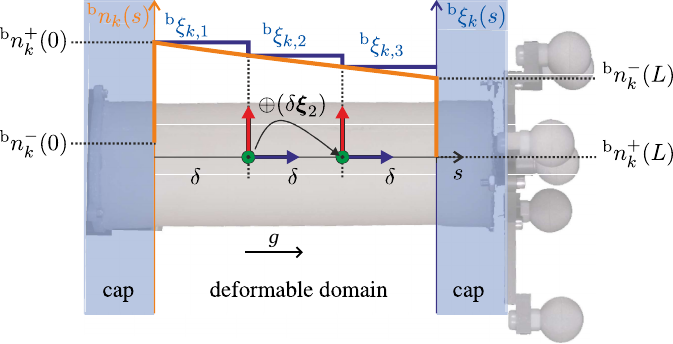}

	\caption{Discretization of the rod model. The illustration shows an example to elaborate on considerations in spatial discretization. 
		Here, two nodes ($\nodes=2$)  as well as three subintervals $n_\mathrm{sub} = 3$ are depicted.
		i) As an example, the steps in scalar internal force $\loq{n}_k$ are shown at $s=0$ and $s=\ell_0$ as an \textcolor{imesorag}{orange line}.
		The index $k$ refers to the $k$-th component of the respective vector quantity.
		The steps introduce the two new state variables and their respective derivative as expressed in \eqref{eq.numerics.transission_condition_discrete}.
		ii) The integration of $\SE$ is conducted for each subinterval with length $\delta$, assuming the strains to be constant in these intervals but linear between two nodes $i$ and $i+1$. 
		This is illustrated with the \textcolor{imesblau}{blue line}.
		The respective algorithm is given in Algorithm~\ref{algorithm.gravity}.
		}
	\label{fig.numerics.computationofstrainsandabsolutepose}
\end{figure}

The second contribution to the external loads $\loq{\externalLoad}$ considered in this work are gravitational loads 
\begin{equation}
	\externalLoad[,g] = - \rho A g \begin{pmatrix} \bs{I}_3 \\ \bs{0}  \end{pmatrix}\,\bs{e}_z
		\label{eq.modelling.distributed_load_gravity_gf}
\end{equation}
with the gravity magnitude $g$ and identity matrix $\identity_3\in\VectorSpace{3\times3}$.
The direction of gravitation is defined to point towards negative $z$ with respect to the inertial frame (compare Fig.~\ref{fig.numerics.computationofstrainsandabsolutepose}). 

For the \ac{lff}, the load is projected into the local frame of the cross-section at $s$ so that
\begin{equation}
	\loq{\externalLoad[,g]}(s,t) = -\rho A g \begin{pmatrix} \identity_3 \\ \zero \end{pmatrix}\, \trans{\rotation}(s,t)\,\bs{e}_z.
	\label{eq.modelling.distributed_load_gravity}
\end{equation}
It follows that for the computation of $\loq{\externalLoad[,g]}$, the absolute orientations are required.
These can be recovered using $\loq{\strain}$ and~\eqref{eq.modelling.rodkinematics}.
Moreover, either iterative algorithms as proposed in Sec.~\ref{sec_collocation_method} or the Magnus expansion \cite{Magnus1954} for linear time-variant matrix \ac{ode} can be applied to compute the integration on $\mathrm{SO}(3)$.
The latter has been done in the context of continuum robotics in the geometrical-variable strain approach \cite{Boyer2021, Mathew2024}.

\subsection{Collocation Method}
\label{sec_collocation_method}
In the first step, an \ac{ocp} of the initial boundary-value problem is defined using the local frame formulation, which consists of \eqref{eq.modelling.lff_compatibility}~and~\eqref{eq.modelling.lff_balance_equations}, in the continuity constraint.
With the boundary values of a cantilevered beam, this leads to
\begin{align}
	\min_{\hlchanges{\bs{x}(s,t)}}       J\left(\stateB(\ell_0),\diff{\stateB(\ell_0)}{t}\right) &\label{eq.numerics.ocp.cost}\\
    \begin{split}
         \subjectto \ 
    \left(\bs{x}(\ell_0) + 
	\begin{pmatrix}
		\bs{0} & \bs{0}\\
		\bs{0} & \dtau
	\end{pmatrix}
	\diff{\bs{x}(\ell_0)}{t}
	\right)
	 & =  \\ \int_{{0}}^{{\ell_0}} \GammaLF(\bs{x}, \diff{\bs{x}}{t}, &\loq{\externalLoad}, \parameterCosserat) \mathrm{d}s 
    \end{split}\label{eq.numerics.ocp.continuity}\\
	\stateA(0,\cdot) &= \diff{\stateA}{t}(0,\cdot) = \bs{0} \label{eq.numerics.ocp.bv}
\end{align}
where the boundary conditions in \eqref{eq.modelling.cosserat.boundaryvalues} are considered in the cost function $J$ and initial condition \eqref{eq.numerics.ocp.bv}.
Note that \eqref{eq.numerics.ocp.continuity} is obtained by rearranging \eqref{eq.modelling.lff_compatibility}~and~\eqref{eq.modelling.lff_balance_equations}, so that the left-hand side of the equation system only contains derivatives \wrt $s$, and then integrating both sides over $s$.
Consequently, the map $\GammaLF$ describes the spatial rate of change of internal stress and velocities with respect to the local frame.
The parameters of $\bs{C}$, $\bs{D}$ and $\bs{M}$ are all contained within $\parameterCosserat$.
For the sake of space, the boundary value for the pose $\htf(0,\cdot) = \identity_{\SE}$ and the initial state $\bs{x}(\cdot, 0) = \bs{0}$ are omitted here.

The implicit midpoint rule
%
is applied to the continuity constraint to approximate the integral, which results in  $\nodes-1$ systems of \acp{ode} $\bs{c}_{\mathrm{eq},i} \in \VectorSpace{12}$.
This method is derived by approximating both $\state$ and its derivative $\diff{\bs{x}}{t}$ with a first-order Lagrange polynomial, \ie~linear functions, and applying the midpoint rule to approximate the integral.
It belongs to the implicit Runge-Kutta methods and is at the same time the simplest collocation method \cite{Hairer1993}(p.~211).

The cost function and the boundary value are discretized too, resulting in the fully implicit differential-algebraic system 

\begin{align}
	\begin{split}
		\bs{c}_{\mathrm{eq},i}(\stated{i},\stated{i+1},\diff{\stated{i}}{t}, \diff{\stated{i+1}}{t},\loq{\externalLoadd[i]}, \loq{\externalLoadd[i+1]}) & = \zero \\
		\with i\in\{0,1,\dots,\nodes-1\}&\\
		\bs{c}_{\mathrm{bv}}(\stated{0}, \stated{\hlchanges{\nodes}},\diff{\stated{0}}{t}, \diff{\stated{\hlchanges{\nodes}}}{t}) &= \zero
	\end{split}
	\label{eq.discrete_nlp}
\end{align}
with discrete states $\stated{i}(t) = \state(s=s_i,t)$, discrete state derivatives $\diff{\stated{i}}{t}(t) = \diff{\state}{t}(s=s_i,t)$ and discretized distributed loads $\externalLoadd[i] = \externalLoad(s = s_i,t)$.

These are input arguments to the implicit \ac{ode} $\trans{\bs{c}_{\mathrm{eq}}} = \inlineMatrix{\trans{\bs{c}_{\mathrm{eq},0}} & \dots & \trans{\bs{c}_{\mathrm{eq},\nodes-1}} }$ and the algebraic equations $\bs{c}_{\mathrm{bv}}$.
The external loads introduce two challenges during discretization that are discussed in the following.

The actuation is introduced via distributed and point loads.
Distributed loads are evaluated by the same implicit midpoint method as the states.
The point loads are generally formulated in \eqref{eq.modelling.cosserat.transitioncondition}, which induces a discrete step function in the spatial domain.
Neglecting the accurate handling of discrete steps in internal loads leads to significant inaccuracy in the integration of strains to obtain the full configuration.
Within the collocation method, this can only be considered accurately at nodes $s_\mathrm{p} \in \{s_0, \dots ,s_{\nodes-1}\}$ that are part of the collocation points and only if additional states are introduced to the system.
A transmission condition including temporal derivatives expresses the point loads
\begin{align}
	\bs{c}_{\mathrm{tc},i_\mathrm{p}} 
    & =  
    \begin{bmatrix}
		\loq{\bs{n}}^+ \\ \loq{\bs{m}}^+
	\end{bmatrix}  
	-  
	\begin{bmatrix}
		\loq{\bs{n}}^- \\ \loq{\bs{m}}^-
	\end{bmatrix}  
	+ (-1)^{i_\mathrm{p}} \loq{\wrenchvector}_\mathrm{p}\\
    \begin{split}
        &= \left(\stateB^{+} + \dtau \diff{\stateB^{+}}{t}\right) - \left(\stateB^{-} +  \dtau \diff{\stateB^{-}}{t}\right) \\
	&+ (-1)^{i_\mathrm{p}} \loq{\wrenchvector}_\mathrm{p} = \zero
    \end{split}
	\label{eq.numerics.transission_condition_discrete}
\end{align}
where $^+$ and $^-$ mark the limit expression at the point $s_\mathrm{p}$ from $\ell_0$ and $0$, respectively, and $i_\mathrm{p} \in \{1,2\}$.
Since the caps (compare Fig.~\ref{fig.numerics.computationofstrainsandabsolutepose}) are not modeled as rods to reduce the number of required nodes, and therefore states, it follows that $s_\mathrm{p,1} = 0$ and $s_\mathrm{p,2} = \ell_0$.
Consequently, twelve new discrete states, $\stateB^{-}(s=0,t)$ and $\stateB^{+}(s=\ell_0,t)$, as well as twelve \acp{ode} are introduced.
The boundary condition then has to be adapted accordingly using $\stateB^+(L,t) = \zero$.
Please refer to Fig.~\ref{fig.numerics.computationofstrainsandabsolutepose} and its caption part `{i)}' for an illustrative elaboration.
For the sake of readability, the new states are simply appended to the original discrete state vector without a change in notation.
This does not affect the validity of the following equations, nor is there any benefit to listing the states separately.

In the next step, the distributed gravitational load in \eqref{eq.modelling.distributed_load_gravity} has to be discretized. 
The required evaluation of distributed gravitational loads $\loq{\externalLoad[,g]}(s, \state)$ could be achieved by a continuous interpolation of the $\SO$ element along the spatial domain, or recursive exponential updates. 
In this work, the latter is considered.
As the states and linearly related strains are assumed to be linear polynomials between two nodes $i$ and $i+1$, the (nonlinear) exponential update is performed in smaller increments of that interval to increase the numerical accuracy.
Each interval $[s_i, s_{i+1})$ is further divided into $n_\mathrm{sub}$ subintervals with length 
\hlchanges{$\delta = \frac{\ell_0}{n_\mathrm{sub}(\nodes-1)}$.}
%
%
The exponential update requires the discrete strains $\loq{\strain}_\mathrm{d}$, which can be computed from the states $\stated{}$ with the affine relation introduced in \eqref{eq.state_x2} using the compliance matrix $\bs{C}$ and pre-curvature $\strain_{0}$.
The discrete strains are computed for the grid $[s_0,\, s_0 + \delta,\, s_0 + 2\delta,\dots,\, s_0 + j\delta,\, \dots,s_{\nodes-1}]$ with $j \in \{0,1,\dots,n_\mathrm{s}-1\}$ by applying the linear interpolation between two states $\stated{i}$ and $\stated{i+1}$.
This is visualized within the example in Fig.~\ref{fig.numerics.computationofstrainsandabsolutepose} and the full algorithm is described in Algorithm~\ref{algorithm.gravity}.
A similar technique has been applied in \cite{Li2023} to discretize the field of strains.

\begin{algorithm}[b]
	\caption{Discrete gravitational load calculation}
	\label{algorithm.gravity}
	\KwIn{$\loq{\strain}_\mathrm{d}$: strain vector, $\parameterCosserat$: parameters}
    \vspace{0.5mm}
	\KwOut{$\loq{\externalLoadd[\mathrm{g}]}$: gravity-load vector}	
	$n_\mathrm{s} \gets$ total number of strain points\;
	$\delta \gets \frac{\ell_0}{n_\mathrm{s}-1}$ \tcp*{Spatial discretization}
	$\htf{}_0 \gets \identity_{\SE}$ \tcp*{Initial orientation}	
	$\bs{\mu}_g \gets \Pmatrix{\identity_3 & \zero} \externalLoad[,g] $ \tcp*{ $\externalLoad[,g]$ is defined in \eqref{eq.modelling.distributed_load_gravity_gf} }
	$\loq{\externalLoadd[\mathrm{g}]}  \gets \externalLoad[,g]$  \tcp*{initial local gravity force}\;	
	\For{$j \gets 1$ \KwTo $n_\mathrm{s}-1$}{
		${\htf}_j \gets {\htf}_{j-1}  \Expm{\loq{\strain}_{\mathrm{d},j} \cdot \delta}{\SE}$\;
		$\rotation_j \gets {\htf}_j.\mathrm{rotation()} \in \SO$\;
		$\loq{\externalLoadd[\mathrm{g},j]} \gets \Pmatrix{\identity_3 \\ \zero} \trans{\rotation}_j  \bs{\mu}_g$\;
	}
	$\loq{\externalLoadd[\mathrm{g}]} \gets \Pmatrix{\loq{\externalLoadd[\mathrm{g},1]} & \dots& \loq{\externalLoadd[\mathrm{g},j]} & \dots & \loq{\externalLoadd[\mathrm{g},n_\mathrm{s}-1]}}$\;
	\KwRet{$\loq{\externalLoadd[\mathrm{g}]}$}
\end{algorithm}

In contrast to the \acp{pde} of the \ac{lff}, the external load due to gravity is spatially integrated using a finer grid for the quadrature.
It follows that the integral within the continuity equation is separated, so that
\begin{align}
	\begin{split}
	&\int\limits_{{0}}^{{\ell_0}} \GammaLF(\bs{x}\st, \diff{\bs{x}\st}{t}, \loq{\externalLoad}\st, \parameterCosserat) \mathrm{d}s\\ 
= &\int\limits_{{0}}^{{\ell_0}} \GammaLF(\bs{x}, \diff{\bs{x}}{t}, \loq{\externalLoad[,\mathrm{p}]}, \parameterCosserat) \mathrm{d}s 
	+ \int\limits_{{0}}^{{\ell_0}}  \loq{\externalLoad[,\mathrm{g}]}(\bs{x},\parameterCosserat) \mathrm{d}s,
	\end{split} \label{eq.numerics.integral_split}
\end{align}
and afterwards, each of these integrals is approximated using different grids.
In this way, the numerical resolution of the right-hand term can be increased without raising the total number of nodes and states. 

All the pieces are merged in the final step of the spatial discretization.
The algebraic equations are inserted into the \ac{ode}, eliminating twelve discrete states (local velocities at the base and local internal wrench at the tip) and their derivatives.
Thereafter, \eqref{eq.discrete_nlp} is rewritten to 
\begin{equation}
		\bs{c}_\mathrm{eq}(\state,\diff{\state}{t}, \bs{u},\parameterCosserat) = \zero
		\label{eq.numerics.implicit_state_space_model}
\end{equation}
where the external loads are omitted, since the system's states fully express these, and the index $\mathrm{d}$ is dropped from now on.
The model inputs $\bs{u}\in\VectorSpace{3}$, here written in the general form known from system theory, are the actuation pressures $\bs{p}$.
This equation is the \textit{state equation in fully implicit form}.

%


\subsection{Explicit State-Space Model and Jacobians}
\label{sec_state_space}
\hlchanges{In the following, the implicit model in \eqref{eq.numerics.implicit_state_space_model} is transformed into an explicit state equation of the form 
\begin{equation}
	\diff{\state}{t} = \bs{f}(\state, \bs{u}, \parameterCosserat).
	\label{eq.numerics.explicit_state_space_model}
\end{equation}}
Due to the simple approach of spatial discretization, this can be done using \ac{cas}.
The implicit model can be expressed as
\begin{equation}
	\tilde{\bs{A}}({\state},\parameterCosserat)\diff{\state}{t} + \tilde{\bs{f}}(\state, \bs{u}, \parameterCosserat) = \zero,
\end{equation}
with $\tilde{\bs{A}} = \diff{\bs{c}_\mathrm{eq}}{\diff{\state}{t}}$ and $\tilde{\bs{f}} = \bs{c}_\mathrm{eq}(\state,\zero,\bs{u},\parameterCosserat)$.
If the inverse $\tilde{\bs{A}}^{-1}$ exists, the equation can be directly rewritten as 
\begin{equation}
	\diff{\state}{t} =  -(\tilde{\bs{A}}({\state},\parameterCosserat))^{-1} \tilde{\bs{f}}(\state, \bs{u}, \parameterCosserat) = \bs{f}(\state, \bs{u}, \parameterCosserat).
	\label{eq.numerics.explicit_state_space_model_defined}
\end{equation}
This form allows for the application of standard numeric solvers to simulate the nonautonomous system.

However, since the initial-value problem is stiff, the choice is limited to implicit solvers, \eg, Matlab's \texttt{ode15s} or \texttt{ode23s}.
These implicit solvers rely on Jacobians of the function $\bs{f}$ \wrt $\state$ (and $t$, if the function is nonautonomous).
If these are not provided, they are computed numerically, which is computationally expensive and therefore undesired.
The derivation of the Jacobian 
\begin{equation}
	\jacobian{\bs{f}}{\state} = \jacobianpd{\bs{f}}{\state} \in \VectorSpace{12N\times12N}
\end{equation}
is mostly straightforward, due to the efficient symbolic implementation of the equations.
The contribution of gravitational load is an exception to that, because of the recursive exponential update in Algorithm~\ref{algorithm.gravity}.
Evaluating this symbolically would lead to large expressions, and, for this reason, the Jacobian is derived analytically using the Lie group framework. 

Considering only gravitational effects on $\diff{\state}{t}$, the Jacobian is broken down into 
\begin{equation}
	\jacobian{\bs{f}_\mathrm{g}}{\state} = 
	\jacobianpd{\bs{f}_\mathrm{g}}{\loq{\externalLoadd[\mathrm{g}]}}
	\jacobianpd{\loq{\externalLoadd[\mathrm{g}]}}{\loq{\strain}_\mathrm{d}}
	\jacobianpd{\loq{\strain}_\mathrm{d}}{\state}
\end{equation}
by applying the chain rule. 
The function $\bs{f}_\mathrm{g}$ is defined by the approximated right-hand term in \eqref{eq.numerics.integral_split}.
It represents the gravitational effects on $\diff{\state}{t}$. 
The crux is to derive the expression $\jacobianpd{\loq{\externalLoadd[\mathrm{g}]}}{\loq{\strain}_\mathrm{d}}$.
Both $\jacobianpd{\bs{f}_\mathrm{g}}{\loq{\externalLoadd[\mathrm{g}]}}$ and $\jacobianpd{\loq{\strain}_\mathrm{d}}{\state}$ are computed via symbolic computation tools.
$\loq{\externalLoadd[\mathrm{g}]}$ is defined in Algorithm~\ref{algorithm.gravity} and consists of the stacked vectors $\loq{\externalLoadd[\mathrm{g},k]}$ with $k = 0 \dots (n_\mathrm{s}-1)$.
Consequently, the task is to find the Jacobian  of the discrete quantity
\begin{align}
	\begin{split}
		    \loq{\externalLoadd[\mathrm{g},k]}
		=    \Pmatrix{\identity \\ \zero}\,
		\trans{\left(\rotation_k(\loq{\strain_\mathrm{d}})\right)} \left(-\bs{\mu}_\mathrm{g}
		\right) \,\with\, {\rotation_0} = \identity_3\\
		\text{and}\, 
		\trans{\rotation_k} = \trans{\left(\rotation_0 \oplus\delta \loq{\angstrain}_{0}  \oplus \dots \oplus \delta{\loq{\angstrain}_{k-1}} \right)} \\
		= \trans{\left(\rotation_0 \Expm{{\delta\loq{\angstrain}}_{0}}{\SO} \dots \Expm{\delta\loq{\angstrain}_{k-1}}{\SO} \right)}      
	\end{split}\label{eq.numerics.discrete_gravitational_load}
\end{align}
with respect to the discrete strains $\loq{\straind[j]}$, where the angular strains $\loq{\angstrain}_{j} = \Pmatrix{\zero_{3\times3} & \identity_{3\times3}} \loq{\straind[j]} $.
This can be addressed by once again applying the chain rule and separating the Jacobian into known elementary Jacobians that can be derived or found in literature \cite{sola_micro_2021}.


This leads to 
\begin{align}
    \bs{\Xi}_{jk} 
	&\coloneqq \trans{\rotation}_k \skw{\bs{\mu}_\mathrm{g}} \Adjoint{\rotation_k}\inv{\Adjoint{\relrotation{j}{k}}} \jacobian{\SO}{\mathrm{r}}(\delta\loq{\angstrain}_{j}),\\
	\jacobianpd{\loq{\externalLoadd[\mathrm{g},k]}}{\loq{\straind[j]}} 
	&= 
	\begin{cases} 
		\Pmatrix{\zero & \bs{\Xi}_{jk}\delta \\ \zero & \zero}, & \text{if } j < k, \\
		\zero, & \text{if } j \geq k.
	\end{cases}
    \label{eq.partial_jacobian}
\end{align}
with $\jacobian{\SO}{\mathrm{r}}$ being the right Jacobian of $\SO$, $j = 0 \dots (n_\mathrm{s}-1)$ and
\begin{align}
	\rotation_k = \relrotation{0}{j-1} \relrotation{j-1}{j} \relrotation{j}{k}, \, \relrotation{j-1}{j} = \Expm{\delta \loq{\angstrain}_{j} }{\SO}.
\end{align}
The Jacobian is then assembled to
\begin{align}
	\jacobianpd{\loq{\externalLoadd[\mathrm{g}]}}{\loq{\strain_\mathrm{d}}} = 
	\begin{bmatrix}
		\zero & \zero & \dots & \zero \\
		\jacobianpd{\loq{\externalLoadd[\mathrm{g},1]}}{\loq{\straind[0]}} & \zero & \dots & \zero \\
		\vdots & \vdots & \ddots & \vdots \\
		\jacobianpd{\loq{\externalLoadd[\mathrm{g},n_\mathrm{s}-1]}}{\loq{\straind[0]}} & \dots & \jacobianpd{\loq{\externalLoadd[\mathrm{g},n_\mathrm{s}-1]}}{\loq{\straind[n_\mathrm{s}-2]}}  & \zero
	\end{bmatrix}.
\end{align}
The evaluation of this part of the full Jacobian is relatively expensive. 
However, the implementation allows for pre-allocation by compiling the gravity functions and their Jacobian.
All derived Jacobians and exported Jacobians are individually verified with their numeric counterpart.

\subsection{Output Models}
\label{sec_output_models}
The full state-space model consists of the state equation and a set of output equations.
In theory, every output that can be described using the states or inputs of the system is applicable.

\begin{figure}[!tb]
\centering
\includegraphics[width=0.56\linewidth]{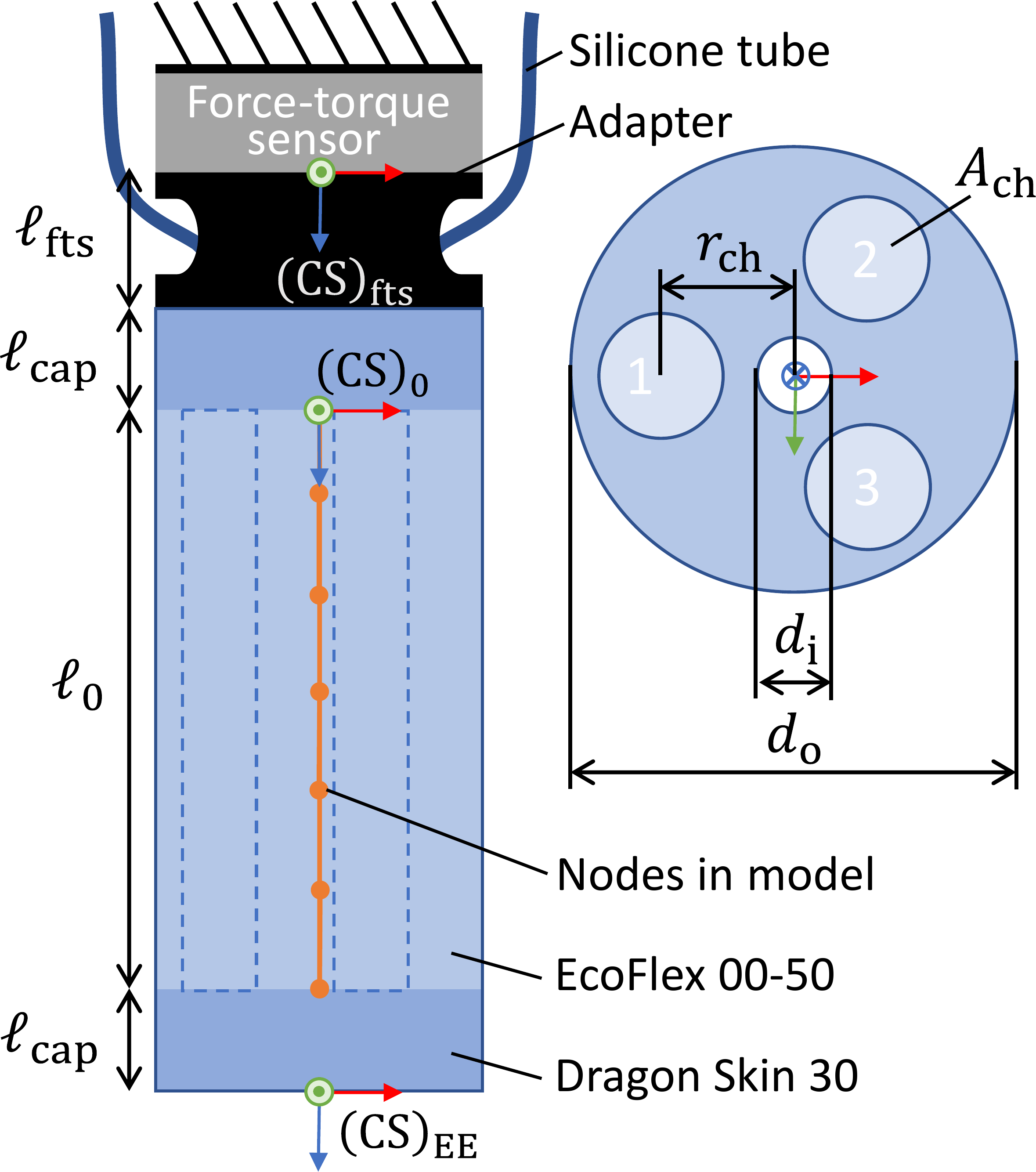}
\begin{tabular}[b]{@{\extracolsep{\fill}} cc}\hline
  Parameter & Value \\ \hline
  $\ell_\mathrm{fts}$ & \SI{0.014}{m} \\
    $\ell_\mathrm{cap}$ & \SI{0.010}{m} \\
  \textcolor{blue}{$\ell_0$} & \textcolor{blue}{\SI{0.125}{m}} \\
  $r_\mathrm{ch}$ & \SI{0.0212}{m} \\
  $A_\mathrm{ch}$ & \SI{235.6}{mm^2} \\
  $d_\mathrm{i}$ & \SI{0.0070}{m} \\
  $d_\mathrm{o}$ & \SI{0.0424}{m} \\ \hline 
\end{tabular}
\caption{Drawing of the soft pneumatic actuator used and a table of the dimensions. The parameter $\ell_0$ written in blue was identified from static data.}
\label{fig_actuator}
\end{figure}

The measured pose of the end effector is described by $\htf[EE]$ \wrt $(\mathrm{CS})_\mathrm{fts}$ (see \figref{fig_actuator}). 
This involves the constant transformations of the 3D-printed interface, the caps manufactured from dragon skin, which are considered to be rigid bodies, and the continuous deformation of the body.
Therefore, the model output is
\hlchanges{
\begin{align}
	\htf[EE] &= \htf[fts](\ell_\mathrm{fts})\htf[cap](\ell_\mathrm{cap})\htf[L](\state)\htf[cap](\ell_\mathrm{cap})\\ 
	\Pmatrix{\bs{y}_\mathrm{pos}\\1 } &= \htf[EE] \Pmatrix{\zero_{3\times1}\\1}, \quad \bs{y}_\mathrm{rot} = \Logm{\rotation_\mathrm{EE}}{\SO} 
\end{align}}
where the tip pose of the rod $\htf[L](\state)$ is computed using Algorithm~\ref{algorithm.gravity}.
The position of the end effector is given by $\bs{y}_\mathrm{pos}$, while the orientation is given by $\bs{y}_\mathrm{rot}$ (parametrized by the tangential space of $\SO$).

\subsection{Exp. 1: Comparison to the Linearized Model}
\label{sec_linerarization}
As already stated, linearized models have been applied in the control of \acp{scr} before \cite{thieffry_control_2019}.
The reason to investigate the linear model here is to underline the requirement of a fully nonlinear system.
The limitations of a linear model
\begin{align}
	\begin{cases}
		\diff{\state}{t} =  
		\evaluated{ \jacobianpd{\bs{f}(\state, \bs{u})}{\state}}{\substack{\state = \state_0 \\ \bs{u} = \bs{u}_0}} (\state - \state_0) + 
		\evaluated{\jacobianpd{\bs{f}(\state, \bs{u})}{\bs{u}}}{\substack{\state = \state_0 \\ \bs{u} = \bs{u}_0}} (\bs{u} - \bs{u}_0), \\ 
		\bs{y}_\mathrm{pos} = 
		\evaluated{\jacobianpd{\bs{h}_\mathrm{pos}(\state)}{\state}}{\state = \state_0} (\state - \state_0) + \bs{h}_\mathrm{pos}({\state_0})
	\end{cases}
	\label{eq.numerics.linear_explicit_state_space_model}
\end{align}
are evaluated in Exp.~1 using a sequence of step functions for the pressure actuation at the equilibrium point $(\state_0, \bs{u}_0)$.
The same input trajectory is repeated for steps with a magnitude \SI{10}{\kilo\pascal} to \SI{80}{\kilo\pascal} in increments of  \SI{10}{\kilo\pascal}.

	
	
	
	

\begin{figure}[b]
	\centering
    \begin{overpic}[]{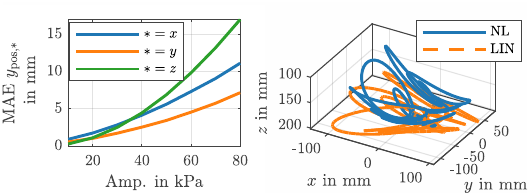}
     \put(0,0){(a)}
     \put(50,0){(b)}
    \end{overpic}
	\caption{\hlchanges{Exp. 1: End-effector positions calculated by the linear (LIN) and nonlinear (NL) model. 
		A trajectory with sequential steps (\SI{80}{\kilo\pascal}) smoothed by a low-pass filter (\SI{4}{\hertz}) is conducted to excite the dynamics.
        (a) \ac{mae} between linear and nonlinear model for the position model outputs over the amplitude of the sequential steps trajectory. 
  	    (b) The trajectory corresponding to  \SI{80}{\kilo\pascal} is displayed.}}
	\label{fig.numerics.linearmodel_comparison}
\end{figure}
%
\hlchanges{
Fig.~\ref{fig.numerics.linearmodel_comparison}(a) shows the \ac{mae} between the nonlinear and linear models for all trajectories, while
Fig.~\ref{fig.numerics.linearmodel_comparison}(b) shows the resulting output trajectory for the maximum magnitude.}
For deflections smaller than \SI{10}{\milli\metre}, a good accordance between linear and nonlinear models is achieved.
However, a super-linear rise of the \acp{mae} can be observed in Fig.~\ref{fig.numerics.linearmodel_comparison}(a) for all translatory \acp{dof}.
The same behavior is observed for the other model outputs that are not displayed here.
This emphasizes the very limited, yet potentially useful, range of applications for the linear model.
\hlchanges{A validation of the nonlinear model using experimental data is provided in \sect{sec_Exp2}.}

%
%
%


\subsection{Soft Pneumatic Actuator and Experimental Setup} \label{sec:SCR+setup}
\begin{figure}[!t]
\centering
\includegraphics[width=\linewidth]{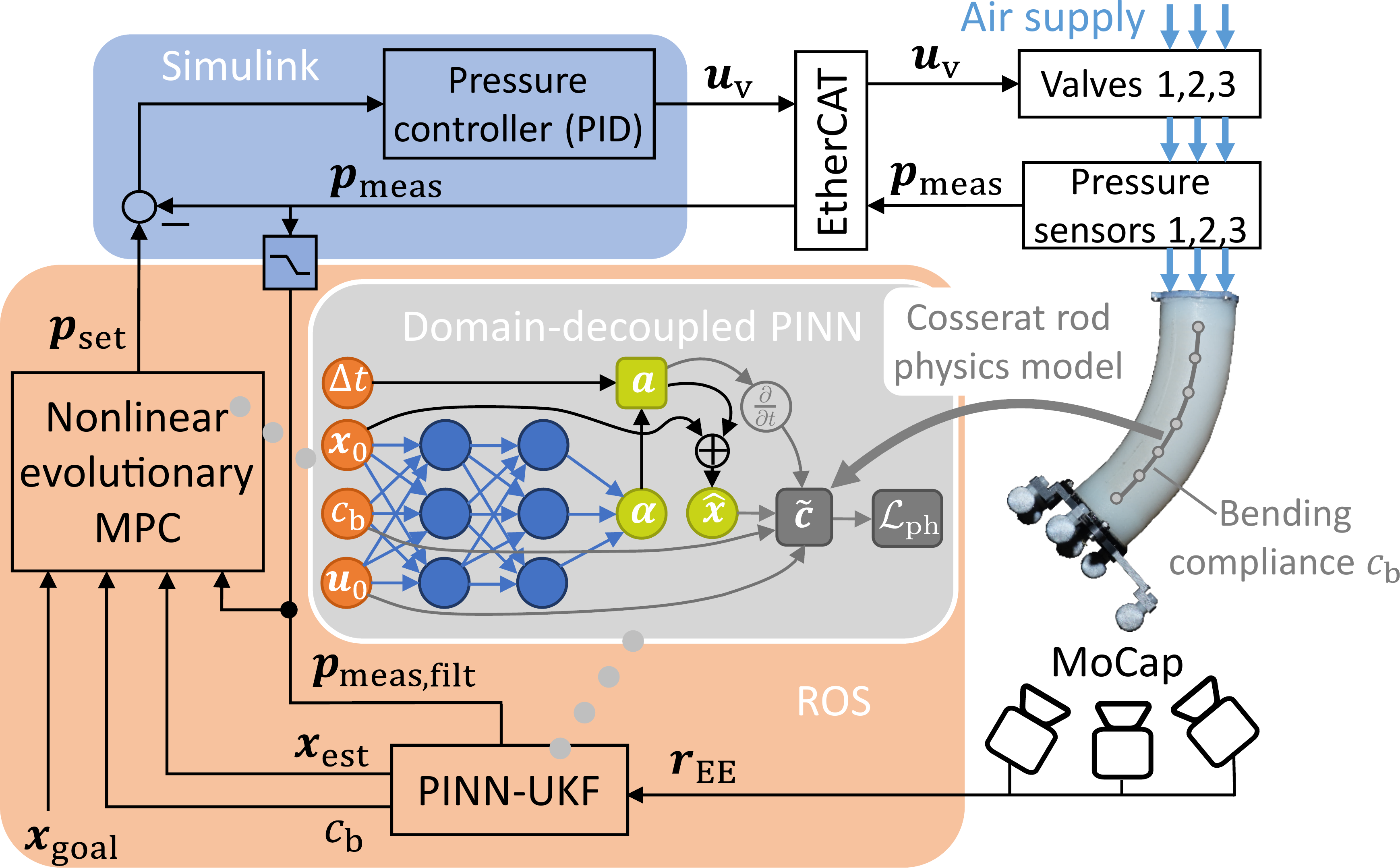}
\caption{Overview of the control architecture including the model predictive control and unscented Kalman filter based on a domain-decoupled physics-informed neural network.}
\label{fig_overview}
\end{figure}

For the experimental validation of the proposed physics-informed neural MPC, we employ a soft pneumatic actuator featuring three fiber-reinforced chambers evenly distributed around its circumference (see \figref{fig_actuator}). The actuator's main body is molded using Smooth-On's EcoFlex 00-50 silicone, while the base and tip caps are fabricated from Smooth-On's Dragon Skin 30 silicone. Further information about the actuator is given in~\cite{bartholdt_parameter_2021}. During the experiments, the actuator is mounted onto an ATI Mini40 force-torque sensor, and motion capture markers are attached to its tip. 
\hlchanges{The pressure in the pneumatic chambers is regulated by 5/3 proportional valves (Enfield, LS-V05s). As shown in \figref{fig_overview}, the valve is controlled at 1 kHz via a PID controller implemented in Simulink, interfaced through EtherCAT I/O modules (Beckhoff EL3164 and EL4004), with pressure feedback provided by sensors (First Sensors, 142BC30A-PCB) placed as close to the chambers as possible.
The inner pressure-control loop, as well as the MPC and UKF, introduced in \sect{sec_pinn_mpc}, are executed on a Linux PC with the PREEMPT\_RT patch (Intel Core i7-14700K, Nvidia GeForce RTX 3070, 32 GB RAM). The PyTorch \ac{nempc} and C++ UKF were implemented as ROS services. 
The motion-capture cameras (OptiTrack, Prime 17W) are connected to a separate PC, which streams pose data over Ethernet (NatNet SDK).
For data logging and live plotting, a dedicated PC connects to the real-time PC via Simulink's external mode.}

\subsection{Parameter Identification} \label{sec_param_ident}

The parameter identification for the Cosserat rod model is carried out in two consecutive stages. First, the static model parameters are estimated using data from 42 stationary configurations that span the actuator’s entire workspace. These configurations include single, dual, and all-chamber actuation, each performed at six discrete pressure levels. A parallelized particle swarm optimization (PSO) algorithm is employed to determine the optimal parameters. The cost function minimizes deviations in the tip position and the reaction forces and moments at the base. To balance these contributions, the \ac{aed} errors for position, force, and torque are weighted by the inverse of the errors from the initial parameter guess. Alternative weighting using the inverse of the respective value ranges was tested, but it resulted in increased dynamic model error.
While most geometric parameters are obtained from the CAD model (see \figref{fig_actuator}), the nominal length $\ell_0$ is identified to allow for corrections (see \tabref{tab:ident_params}). 
The mass of the tip cap and marker holder, denoted as $m_\ind{EE}$, is represented by an increased density between the last two nodes in the simulation. The rigid body at the tip is currently not modeled to reduce the number of states. This simplification likely deviates from the true center of mass, and therefore, $m_\ind{EE}$ is treated as an identification parameter to minimize the model error. 
In addition, the bending compliance $c_\ind{b}$ and dilatation compliance $c_\ind{dilat}$ are identified. Preliminary investigations indicated that the remaining compliance parameters are not sufficiently excited by the available measurement data and cost function. Consequently, these parameters are estimated analytically based on the material properties. 
Following the static parameter identification, the dynamic parameter was identified using a dynamic trajectory (\SI{10}{s}). This trajectory consists of periodic oscillations at frequencies of \SI{0.1}{Hz}, \SI{0.5}{Hz}, and \SI{1}{Hz}, followed by a step to ambient pressure. The same weighting factors applied in the static identification were used to normalize the position, force, and torque cost.
The dynamic parameter identified is the bending damping $d_\ind{b}$ (see \tabref{tab:ident_params}).  Due to the direct identification of the diagonal elements of $\boldsymbol{D}_\tau$, rather than $\boldsymbol{D}$ (see \eqref{eq_material_law}), the resulting unit is rather unusual. The remaining damping parameters could not be reliably identified due to insufficient excitation in the available measurement data and were estimated analytically based on the material properties and geometry.
A comparison of the Cosserat model using the identified parameters to the measurements is provided in Exp.~2 (\sect{sec_Exp2}).

\begin{table}[!tb]
    \centering
    \caption{Identified model parameters}
    \label{tab:ident_params}
    \footnotesize
    \setlength{\tabcolsep}{4pt}
    \renewcommand{\arraystretch}{1.1}
    \begin{tabular}{@{}c|c|c|c?c@{}}
        \hline
        \multicolumn{4}{c?}{\textbf{Static}} & \textbf{Dynamic} \\ \hline
        $\ell_0$ (m) & $m_\mathrm{EE}$ (kg) & $c_\mathrm{b}$ ($\mathrm{rad/(N\,m^2)}$) & $c_\mathrm{dilat}$ ($\mathrm{m/(N\,m)}$) & $d_\mathrm{bend}$ (s) \\ \hline
        0.1249 & 0.0668 & 49.9 & 0.0059 & 0.005 \\ \hline
    \end{tabular}
\end{table}

 Preliminary experiments revealed a short delay between the set and measured pressures of the pressure controller \hlchanges{(see \sect{sec:SCR+setup})}. To consider this delay in the MPC, the actuation dynamics were identified using MATLAB’s \texttt{tfest} function. A first-order low-pass filter provided the best fit  

\begin{equation}
    G_\ind{p}(s) = \frac{11}{s+10.94}
\end{equation}
$s$ denotes the frequency parameter in the frequency domain here. For use in the MPC, the continuous transfer function was discretized using MATLAB’s \texttt{c2d} function and implemented in Python using SciPy’s \texttt{lfilter} function.


\section{Domain-decoupled Physics-informed Neural Networks for Surrogate Modelling} \label{sec_pinn}
Training \acp{pinn} with physics-based loss functions enables them to generalize beyond the experimental data used during training. As a result, the amount of required experimental data can be significantly reduced. When the underlying physics model is sufficiently accurate, it is even possible to eliminate the need for experimental training data entirely~\cite{habich_generalizable_2025}. 
Before explaining our implementation and PINN training in \sect{sec_pinn_training} and prediction results in \sect{sec_Exp2}, the \acf{ddpinn} presented in \cite{krauss_domaindecoupled_2024} is introduced (\sect{sec_ddpinn}).

\subsection{Domain-Decoupled PINN} \label{sec_ddpinn}
The \ac{ddpinn}~\cite{krauss_domaindecoupled_2024} employs time-dependent ansatz functions to improve training speed and prediction accuracy. \acp{ddpinn} estimate the future state $\hat{\bs{x}}$ after a sampling time interval $t_\ind{s}$ based on the current state $\bs{x}_0$ and the current control input $\bs{u}_0$, using ansatz functions $\bs{a}$
\begin{equation}\label{eq_pinn_predict}
    \hat{\bs{x}}(t_\ind{s})=\bs{f}_\ind{PINN}(t_\ind{s},\bs{x}_0,\bs{u}_0)=\bs{x}_0+\bs{a}(\bs{\alpha}(\bs{x}_0,\bs{u}_0),t_\ind{s}).
\end{equation}
The resulting architecture (with an arbitrary number of hidden nodes) is shown in \figref{fig_overview}. 
The use of time-dependent ansatz functions facilitates learning the dynamics of more complex systems by enabling a more efficient parameterization of the solution space. 
Further, they enable training with analytic gradients (instead of automatic differentiation). 
Krauss et al.~\cite{krauss_domaindecoupled_2024} show that this leads to faster training compared to other \acp{pinn} for control, such as those proposed in~\cite{antonelo_physicsinformed_2024} or \cite{nicodemus_physicsinformed_2022}, for which training time scales overproportionally with the number of system states. 
While the choice of ansatz functions is critical for an accurate system approximation, the ansatz function defined in~\eqref{eq_ansatz} has  performed well across a variety of systems~\cite{krauss_domaindecoupled_2024, habich_generalizable_2025} and will also be used in this work:
\begin{multline}\label{eq_ansatz}
\bs{a}(\bs{\alpha},t){=}\sum_{i=1}^{n_\ind{a}} \bs{\alpha}_{1i}{\odot}\big(\sin{(\bs{\alpha}_{2i}t{+}\bs{\alpha}_{3i})}{\odot}\exp{({-}\bs{\alpha}_{4i}t)}\\[-6pt]
{-}\sin{(\bs{\alpha}_{3i})} \big).
\end{multline}
Incorporating the element-wise $\sin(\cdot)$ and $\exp(\cdot)$ functions, this ansatz function resembles a sum of damped oscillations. $\odot$ represents the element-wise Hadamard product and $n_\ind{a}$ denotes the number of ansatz functions. By using this function, the solution space is parameterized using the ansatz tensor 
\begin{equation}
    \bs{\alpha} = \trans{[\bs{\trans{\alpha}_1},\trans{\bs{\alpha}_2},\trans{\bs{\alpha}_3},\trans{\bs{\alpha}_4}]} \in \mathbb{R}^{4\times\ind{dim}(\bs{x})\times n_\ind{a}}.
\end{equation}
Because $\bs{a}(\bs{\alpha},0){\equiv}0$, the initial condition $\bs{x}_0 = \bs{f}_\ind{PINN}(0,\bs{x}_0,\bs{u}_0)$ is automatically satisfied when using this ansatz function. The model training and the optimization of hyperparameters are described in \sect{sec_pinn_training}.  The code for the model training is published open-source\footnote{\url{https://github.com/sniklasbarth/rod-pinn}}.

\subsection{PINN Training and Hyperparameter Optimization} \label{sec_pinn_training}
As a trade-off between modeling accuracy, complexity, and computational cost, the Cosserat rod was discretized into $\nodes=7$ nodes for training the \ac{pinn}, resulting in a total of $n_\ind{x} = 72$ states. Previous investigations have shown that the bending stiffness varies across the workspace~\cite{bartholdt_parameter_2021} and that adaptive bending stiffness can therefore be leveraged to minimize the modeling inaccuracies~\cite{mehl_adaptive_2024}. Consequently, adaptive state and parameter estimation are essential for achieving precise control. To utilize the \ac{pinn} within this application, the bending compliance must be treated as an input to the network, and the \ac{pinn} has to be trained across varying bending stiffness values. Previous work~\cite{habich_generalizable_2025} has shown that the \ac{ddpinn} architecture is capable of this parameter adaptivity. Although the training of \acp{ddpinn} using experimental data is feasible, we opted to train exclusively on the physics-based model. This choice facilitates training with variable bending compliance, as the physics model can simulate different stiffnesses consistently, whereas the stiffness in real-world experiments is not explicitly known.
Hence, the prediction equation \eqref{eq_pinn_predict} is modified to
\begin{equation}\label{eq_pinn_predict_mod}
    \hat{\bs{x}}(t_\ind{s})=\Tilde{\bs{f}}_\ind{PINN}(t_\ind{s},\bs{x}_0,\bs{u}_0,{c}_\ind{b}).
\end{equation}
A simple simulated trajectory was used to determine the viable range of states, which is used to normalize the states and control inputs to a range of $[-1,1]$ and to define the sampling space for the collocation points. During one training epoch, $10^6$ collocation points were evaluated in batches of $n_\ind{batch} = 1000$~points. For the states, these are sampled from a normal distribution with a standard deviation of $\sigma = 0.5$, while for the inputs and the bending compliance, Latin hypercube sampling was used. The physics loss is implemented using the implicit state equation \eqref{eq.numerics.implicit_state_space_model} with the bending compliance as an additional input. The next state $\hat{\bs{x}}$ as the output of the \ac{ddpinn} and the altered bending compliance $\Tilde{c}_\ind{b}$ are inserted into the modified implicit state-space model $\Tilde{\bs{c}}_j(\hat{\bs{x}},\bs{x}_0,t_\ind{s},\bs{u}_0,\Tilde{c}_\ind{b})$ to calculate the physics loss
\begin{equation} \label{eq_physics_loss}
\mathcal{L}_\mathrm{ph} = \sum_{i=1}^{n_\ind{batch}} \left[ \frac{1}{n_\ind{x}}\sum_{j=1}^{n_\ind{x}} \Tilde{\bs{c}}_j(\hat{\bs{x}},\bs{x}_0,t_\ind{s},\bs{u}_0,\Tilde{c}_\ind{b})^2 \right]_i.
\end{equation}
To obtain the inputs of \eqref{eq.numerics.implicit_state_space_model} from the inputs shown in \eqref{eq_physics_loss}, the state's time derivative $\dot{\bs{x}}$ is calculated using the analytic differentiation of the ansatz function, while the gravitational and pneumatic loads $\bs{\zeta}_\ind{ext,g},\bs{\zeta}_\ind{ext,p}$ are calculated using the respective functions of the physics model from \sect{sec_cosserat_model}.

For training in PyTorch, the optimizer ADAM was used in combination with learning rate scheduling, employing a patience of $75$~epochs and a reduction factor of $0.5$. 
Prior to hyperparameter optimization, various sampling frequencies in the range of $[50, 200]$\,Hz were tested to identify the minimum frequency that still ensures good convergence. A high sampling time is desirable, as the computation time of the MPC must remain below the sampling time to enable real-time operation. Additionally, it benefits the prediction of long time horizons with fewer prediction steps. In consideration of these factors, a sampling frequency of \SI{70}{Hz} was selected. 

\begin{figure}[!tb]
\centering
\includegraphics[width=0.8\linewidth]{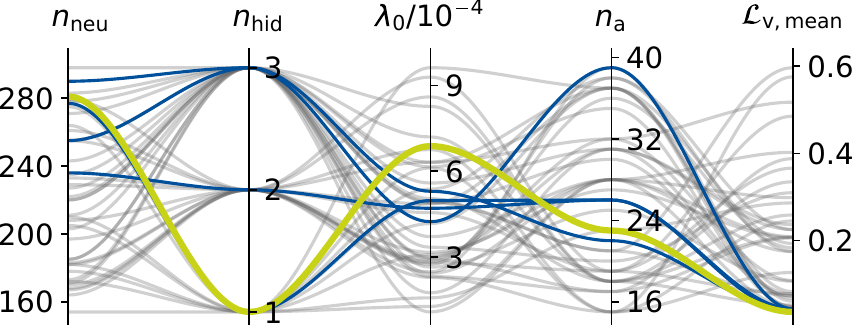}
\caption{Results of the hyperparameter optimization for \ac{ddpinn} training. The best parameter set is shown in green. The next four best parameter sets are plotted in blue. Parameter sets with  $\mathcal{L}_\ind{v,mean}>100$ are excluded.}
\label{fig_hpo}
\end{figure}

With the sampling frequency fixed, a hyperparameter optimization was conducted using the asynchronous successive halving algorithm (ASHA)~\cite{li_system_2020a} to determine the most effective network architecture. The parameters optimized include the number of neurons per layer $n_\ind{neu}$, the number of hidden layers $n_\ind{hid}$, the initial learning rate $\lambda_0$, and the number of ansatz functions $n_\ind{a}$. A total of 72 neural networks with different configurations were trained in parallel. Based on their validation loss, some networks were terminated early after a grace period of 300 epochs, while the majority were trained for 1500 epochs.
The results of this optimization are visualized in \figref{fig_hpo}. The parameter set yielding the lowest mean validation loss $\mathcal{L}_\ind{v,mean}$ (highlighted in green) was $\{n_\ind{neu}=281, n_\ind{hid}=1, \lambda_0=0.000688, n_\ind{a}=23 \}$. This best-performing network was subsequently trained for an additional 1500 epochs to ensure complete convergence. The total training time for this network was slightly over eight days using two Intel Xeon Gold 6442Y 2.6 GHz CPU cores and 8 GB RAM.

\subsection{Exp. 2: Model Validation and PINN Verification} \label{sec_Exp2}
\begin{figure}[!tb]
\centering
\includegraphics[width=\linewidth]{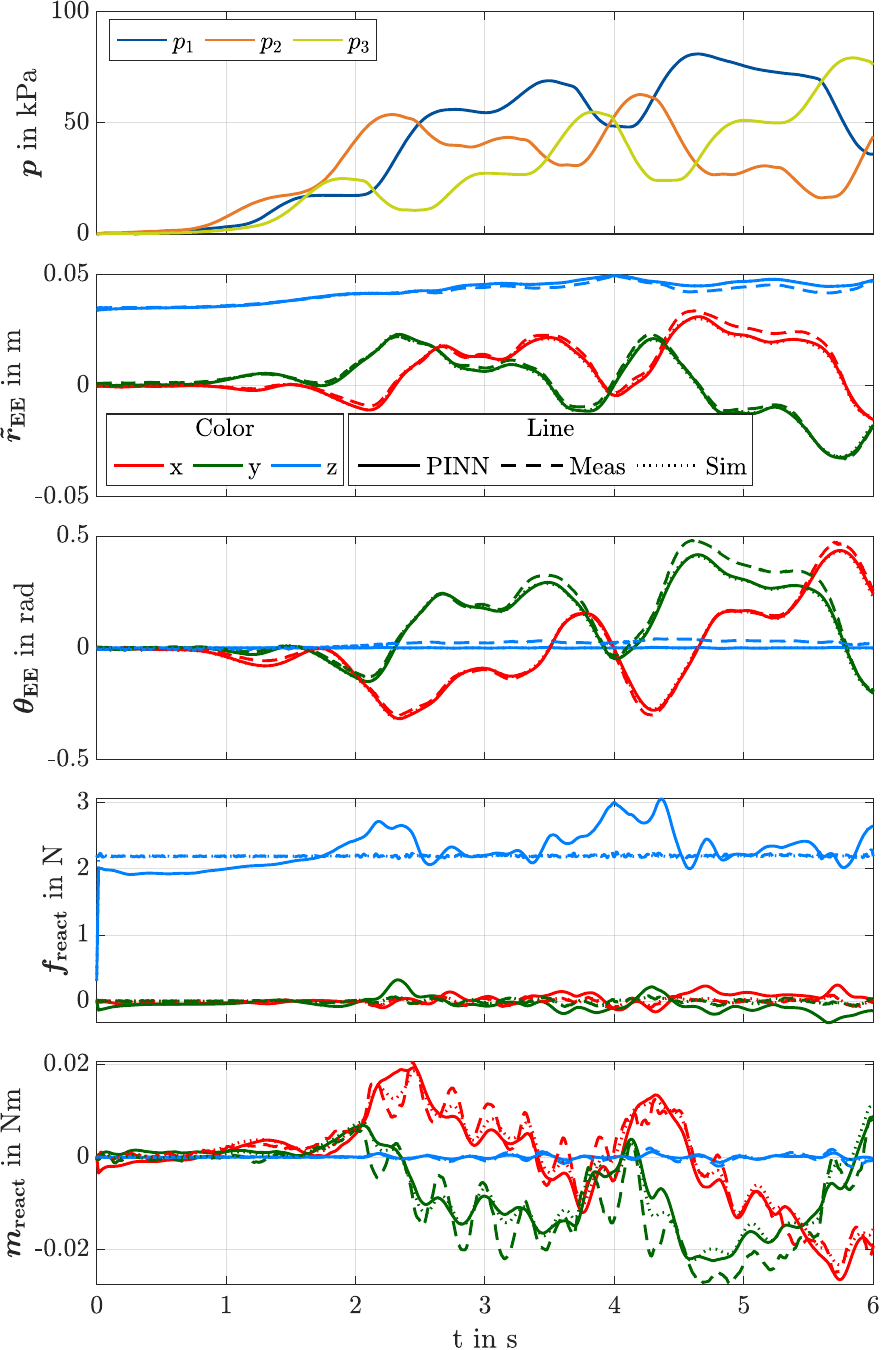}
\caption{Exp. 2: Comparison of the end-effector position and orientation as well as base reaction force and torque predicted by the \ac{pinn} and the first-principle simulation to real measurements. $\bs{\theta}_\ind{EE}$ represents the tangent space of the SO(3) group (equivalent to the rotation vector of the axis-angle representation).}
\label{fig_PINN_predict}
\end{figure}

To validate the first-principles model, denoted as ``Sim", and verify the \ac{ddpinn}, the predictions of the first-principles simulation and \ac{ddpinn} are plotted with pose and wrench measurements, denoted as ``Meas", for a dynamic input trajectory in \figref{fig_PINN_predict}. The measurements are captured using the experimental setup described in \sect{sec:SCR+setup}. The measured pressures were filtered using a second-order Butterworth filter with a cutoff frequency of \SI{10}{Hz} \hlchanges{to reduce high-frequency measurement noise} and then applied as input to the physics-based model and \ac{ddpinn}.
The modified end-effector position $\bs{\Tilde{r}}_\ind{EE} = \bs{r}_\ind{EE}-\trans{[0,0,\ell_0]}$ is shown to enhance the visibility of deviations when visualizing all dimensions in one plot.

The validation errors (\ac{aed}) of the first-principles simulation from \sect{sec_cosserat_model} \hlchanges{are listed in \tabref{tab:val_errrors}. The end-effector position error is equivalent to \SI{1.8}{\percent} of the actuator's length.}

\begin{table}[!tb]
    \centering
    \caption{Simulation-validation errors and \ac{pinn}-verification errors}
    \addtolength{\tabcolsep}{-0.2em}
    \renewcommand{\arraystretch}{1.3}
    \begin{tabular}{l l|c|c|c|c}
        &&$\bs{r}_\ind{EE}$& $\bs{\theta}_\ind{EE}$ & $\bs{f}_\ind{react}$ & $\bs{m}_\ind{react}$  \\
        &&mm & rad & N & Nmm \\
        \hline
        \multirow{2}{*}{\shortstack[c]{Simulation-\\measurement error}} & Mean &2.42 & 0.0382 & 0.026 & 3.46 \\ \cline{2-6}
        & SD &1.52 & 0.0247 & 0.092 & 2.61  \\ \hline
        \multirow{2}{*}{\shortstack[c]{PINN-simulation\\error}} & Mean & 0.549 & 0.00630 & 0.259 & 2.03  \\ \cline{2-6}
        & SD & 0.377 & 0.00433 & 0.177 & 1.18  \\ \hline
    \end{tabular}
    \label{tab:val_errrors}
\end{table}

Hence, the first-principles model prediction of the position, orientation, and force data shows good agreement with the measured signals, but \hlchanges{the force error has a comparatively high variance and} the dynamic behavior of the torque output does not match well.
This is most likely due to a simplification of the inertia of the tracking system, which is implicitly incorporated as an increased density of the material towards the rod's end. \hlchanges{This causes deviations when the tip is accelerated, but the} shortcut helps to reduce the number of states and the computation time, which is important for real-time control. 
For the end-effector position and orientation, as well as the base reaction moment, the approximation error between the \ac{ddpinn} and the first-principles simulation\hlchanges{, listed in \tabref{tab:val_errrors},} is smaller than the error of the physics simulation.

\begin{figure*}[!tb]
\centering
\includegraphics[width=\linewidth]{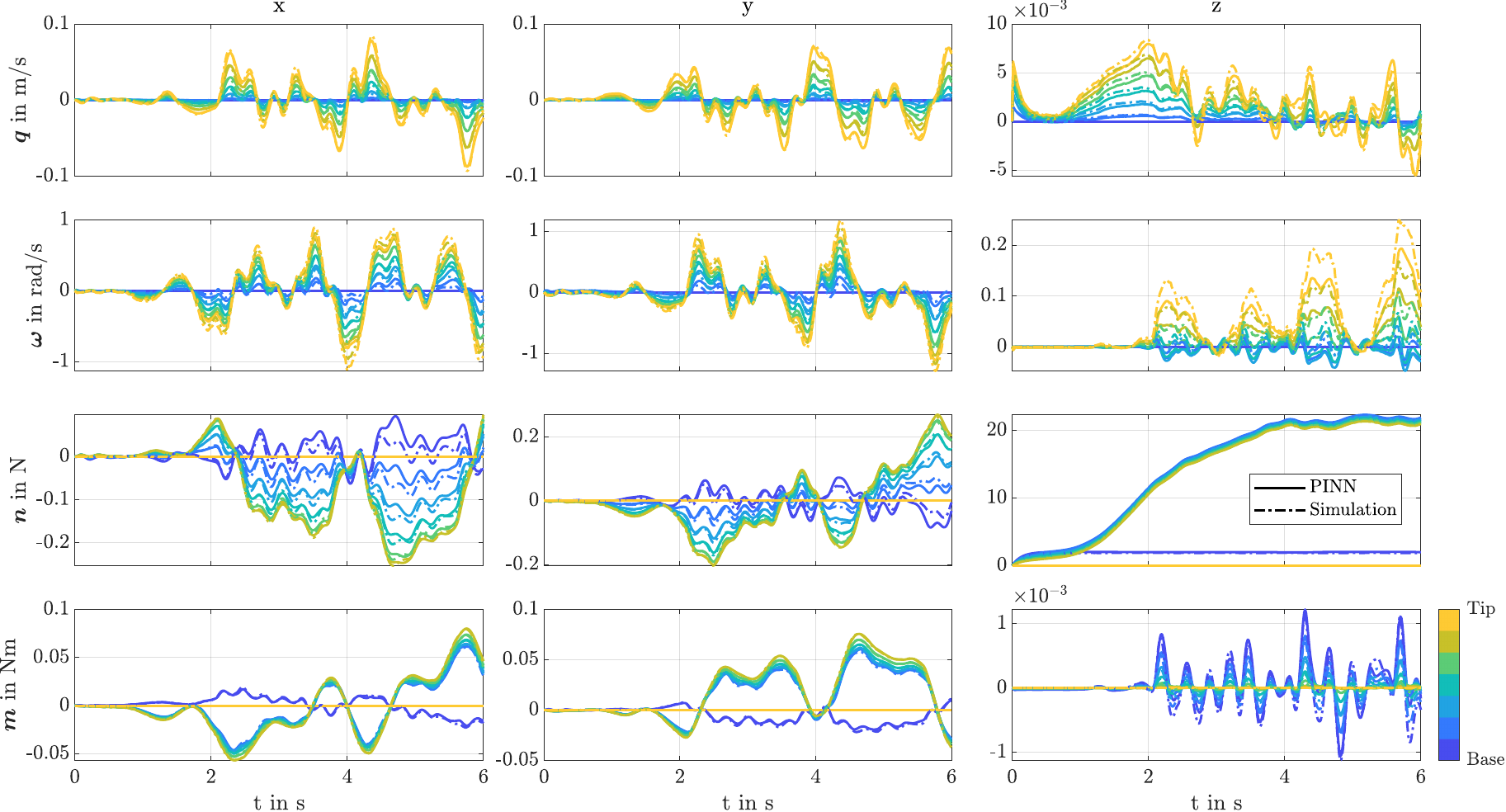}
\caption{Exp. 2: Comparison of the states predicted by the PINN and the first-principles simulation. The used pressure trajectory is shown in \figref{fig_PINN_predict}.}
\label{fig_PINN_veri}
\end{figure*}

However, for the base-reaction force, the error between the \ac{ddpinn} and the simulation is significantly larger than the simulation-to-measurement error.
This suggests that while the \ac{pinn} is generally capable of approximating the physics simulation accurately, its prediction accuracy varies across different states. Looking at the predicted states of the \ac{pinn} and physics simulation in \figref{fig_PINN_veri}, this can be verified. Here, the linear velocities $\bs{q}$, angular velocities $\bs{\omega}$, internal forces $\bs{n}$, and internal moments $\bs{m}$ in the local x, y, and z directions are plotted. While for most states the deviations are almost invisible, a \hlchanges{considerable relative} deviation is visible for the angular velocities in the z direction and the internal forces in the x and y directions. These deviations presumably result in the deviation of the base-reaction force.

\begin{table}
    \centering
    \renewcommand{\arraystretch}{1.3}
    \caption{Mean computation times of the model prediction and end-to-end controller update per step (\SI{14.28}{ms}) for different computation approaches (averaged over 1000 steps)}
    \begin{tabular}{l|l}
        \hline
        Physics simulation using \texttt{ode23s} on CPU (Matlab) & \SI{35.9}{ms}\\ \hline
        DD-PINN with $n_\ind{batch}=1$ on CPU (Matlab) & \SI{1.23}{ms}\\ \hline
        DD-PINN with $n_\ind{batch}=147$ (UKF) on GPU (PyTorch) & \SI{3.30}{\micro \second}\\ \hline
        DD-PINN with $n_\ind{batch}=1000$ (NEMPC) on GPU (PyTorch) & \SI{0.802}{\micro \second}\\ \hline\hline  
        \hlchangestwo{End-to-end physics UKF + MPC on CPU (Matlab)*} & \hlchangestwo{\SI{157}{\second}}\\ \hline 
         \hlchangestwo{End-to-end PINN UKF + MPC on GPU (PyTorch + Eigen)} & \hlchangestwo{\SI{12.3}{\milli \second}}\\ \hline 
        \multicolumn{2}{c}{\scriptsize \hlchangestwo{*due to the large computation time only 200 steps have been used for averaging}} \\
    \end{tabular}
    \label{tab:comp_times}
\end{table}

Additionally, the computation times required for predicting one timestep (\SI{14.28}{ms}; corresponds to the control frequency of \SI{70}{Hz} in \sect{sec_NEMPC}) are compared for the first-principles simulation and the \ac{pinn} with various computation approaches (see Table~\ref{tab:comp_times}).
\hlchanges{A trajectory (1000 steps) with a course similar to the one in \figref{fig_PINN_veri} was used to evaluate the computation times} on a Windows laptop equipped with an AMD Ryzen 9 8945HS 4.00\,GHz CPU, an Nvidia GeForce RTX 4070 Laptop GPU, and 32\,GB RAM.
For the first-principles simulation, an \texttt{ode23s} solver was used in Matlab as it yielded the most accurate results of the tested solvers.
It is partially optimized for speed using compiled functions.
For single evaluations, a speed-up factor of 29.2 can be achieved by the \ac{pinn}.
For batch-wise evaluation, which is used in the UKF and NEMPC algorithms (\sect{sec_pinn_mpc}),
\hlchangesthree{the times in Table~\ref{tab:comp_times} are given per sample, i.e., the measured wall-clock time of one batched forward pass (including GPU synchronization) divided by $n_\ind{batch}$.
The physics simulation was evaluated sequentially for a single state, so the resulting speed-up factors of \SI{10900}{} ($n_\ind{batch}=147$) and \SI{44800}{} ($n_\ind{batch}=1000$) are partly caused by parallel evaluation of many samples.
Hence, batched and GPU-optimized ODE solvers would offer a more direct comparison. However,} \hlchanges{existing PyTorch-based libraries lack the adaptive (semi-)implicit schemes required for the stiff dynamics of Cosserat rods.
While specialized frameworks like \texttt{diffrax} (JAX) or \texttt{DifferentialEquations.jl} (Julia) provide these capabilities, adaptive implicit solvers are fundamentally poorly suited for GPU parallelization (Single Instruction, Multiple Threads).
Their non-deterministic execution flow causes thread divergence~\cite{niemeyer_accelerating_2014}, whereas PINNs leverage deterministic matrix operations, which GPUs are made for.}

While \hlchanges{matching the physics model's accuracy}, the \ac{pinn} can speed up the prediction by a factor of up to \SI{44800}{} compared to the physics simulation.
Compiling the \ac{pinn} using TensorRT or TorchScript did not accelerate computation.
However, in general, the computation times are only approximate values because the computation time is affected by many factors.

\section{Model-Predictive Control of SCRs Using Physics-informed Surrogates} \label{sec_pinn_mpc}
To realize closed-loop control \hlchanges{of the Cosserat rod state $\state$} using the \ac{ddpinn} described above, a control algorithm that can handle the underactuation and a state-estimation approach that estimates the model states from available measurements are needed. This section first describes the \acf{ukf} (in \sect{sec_UKF_design}) used for state estimation and proves its parameter estimation capabilities in numeric simulations. Second, the \acf{nempc} algorithm is introduced, and the control results in simulation and the real world are shown (\sect{sec_NEMPC}).

\subsection{Unscented Kalman Filter for State Estimation} \label{sec_UKF_design}
The \acf{ukf}~\cite{julier_unscented_2004} is a well-established method for state estimation in nonlinear systems. It leverages weighted sigma points, generated via the unscented transform, to approximate the distribution of the state without requiring linearization. This enables effective handling of nonlinear dynamics while maintaining significantly lower computational complexity compared to particle filters.
To enable simultaneous parameter estimation, specifically of the bending compliance, this model parameter is treated as an augmented state. It remains unchanged in the state transition model but is updated based on observations. 

To achieve real-time capability, the \ac{ukf} in this work employs the \ac{ddpinn} as a surrogate model for the dynamic Cosserat rod equations, significantly reducing computation time while preserving model fidelity.
Initially, the \ac{ukf} was implemented in MATLAB for simulation and parameter tuning. 
For the real-time implementation at the test bench, a UKF ROS node was developed in C++. 
The observation function was code-generated from Matlab and compiled into a shared C++ library, while the traced PyTorch model was integrated using LibTorch. 
The UKF algorithm itself was implemented using an open-source repository~\cite{dangio_pcdangio_2024}.

\subsubsection{UKF Parameter Tuning}
Optimization of the \ac{ukf} parameters was performed using parallel \acf{pso} based on a dynamic trajectory. Since no ground truth of the system state is available, the cost function was designed to reflect observable errors and undesirable filter behavior.
Specifically, the cost function includes the error between the estimated and measured end-effector position and orientation ($e_\ind{EE}$, $e_\ind{ori}$). 
To encourage physically plausible shapes, the smoothness of the strain distribution along the arc length is evaluated by computing the moving standard deviation of the strain derivatives with respect to $s$, penalizing non-smooth deformations ($e_\ind{kink}$).
Additionally, the cost function penalizes the occurrence of NaN values (counted as $n_\ind{NaNs}$) in the estimated state vector, resulting from the \ac{ddpinn} and divergence of the \ac{ukf}, defined as the loss of positive definiteness in the covariance matrix ($e_\ind{div}$). These factors are included in the cost function 
\begin{equation}
    {f}_\ind{cost} = w_\ind{EE} e_\ind{EE} + w_\ind{ori} e_\ind{ori} + w_\ind{kink} e_\ind{kink} + w_\ind{NaNs} n_\ind{NaNs} + e_\ind{div}
\end{equation}
as additive penalties to ensure numerical stability and physical consistency of the estimated states.
The weights in the cost function are chosen as the inverse of the respective initial errors to balance the influence of each term.
In addition to the \ac{ukf} parameters $\alpha$, $\beta$, and $\kappa$, both the measurement noise $R_{i,i}$ and the process noise covariances $q_i$ are tuned using the \ac{pso}. To limit the number of optimization parameters, the same noise level is assumed for all spatial directions and nodes within each state group. Specifically, a single parameter is used for internal forces in x and y, internal forces in the z direction (different because of gravity), internal moments, linear velocities, and angular velocities, respectively. One additional parameter is tuned for the bending compliance process noise (see Table~\ref{tab:ukf_param}).
\begin{table}[h]
    \centering
    \renewcommand{\arraystretch}{1.3}
    \caption{UKF parameters optimized using a PSO.}
    \begin{tabular}{ccccc}
        \hline
         $q_\ind{velo}$ & $q_\ind{angvel}$ & $q_\ind{force}$ & $q_\ind{forceZ}$ & $q_\ind{torque}$ \\ \hline
         $8.92\cdot 10^{-6}$ & $2.76\cdot 10^{-5}$ & $1.26\cdot 10^{-5}$ & $0.0436$ & $1.49\cdot 10^{-7}$  \\ \hline
         $q_\ind{cbend}$ & ${R}_{i,i}$ & $\alpha$ & $\beta$ & $\kappa$  \\ \hline
         $1.11\cdot 10^{-5}$ & $2.59\cdot 10^{-8}$ & 0.1 & 1.097 & 10.13 \\ \hline
    \end{tabular}
    \label{tab:ukf_param}
\end{table}

\subsubsection{Exp. 3: Bending compliance parameter estimation} \label{sec:Exp3}
\begin{figure}[!tb]
\centering
\includegraphics[width=.7\linewidth]{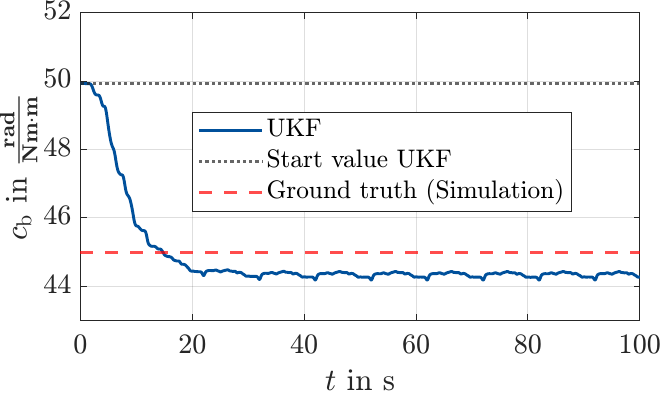}
\caption{Exp. 3: Estimation of the bending compliance $c_\ind{b}$ in simulation.}
\label{fig_ukf_param_est}
\end{figure}

In \figref{fig_ukf_param_est}, the \ac{ukf} was applied to simulated data (dynamic trajectory consisting of three superposed oscillations up to \SI{0.8}{Hz}) with $c_\ind{b} = 45\,\frac{\text{rad}}{\text{Nm}\cdot\text{m}}$, while the initial bending compliance parameter in the \ac{ukf} was set to the identified value of $c_\ind{b,ident} = 49.9\,\frac{\text{rad}}{\text{Nm}\cdot\text{m}}$. This setup allows evaluation of the UKF's parameter estimation capabilities. The \ac{ukf} converges towards the true bending compliance. After approximately \SI{20}{s}, it estimates a bending compliance of \SI{44.45}{\frac{\text{rad}}{\text{Nm}\cdot\text{m}}} \hlchanges{(mean over $[20,100]$\,s)} with very little oscillation, which corresponds to a relative error of \SI{1.2}{\percent}.

This proves the \ac{ukf}'s parameter-estimation capability
.
The residual error after \SI{20}{s} likely stems from mismatch between the physics model and the \ac{pinn}. Higher process noise on the bending-compliance state could speed up convergence but risks instability from parameter drift.

\subsection{Nonlinear Evolutionary Model-Predictive Control} \label{sec_NEMPC}
\ac{mpc} is a widely adopted optimal control approach known for its strong real-world performance. At each time step, it optimizes a control trajectory over a finite horizon from the current state, applies only the first input, and repeats. Most implementations use gradient-based solvers to find a locally optimal solution that satisfies the constraints, but these often cannot meet real-time requirements for systems with complex nonlinear dynamics~\cite{hyatt_realtime_2020}.


\begin{figure}[!tb]
\centering
\includegraphics[width=0.7\linewidth]{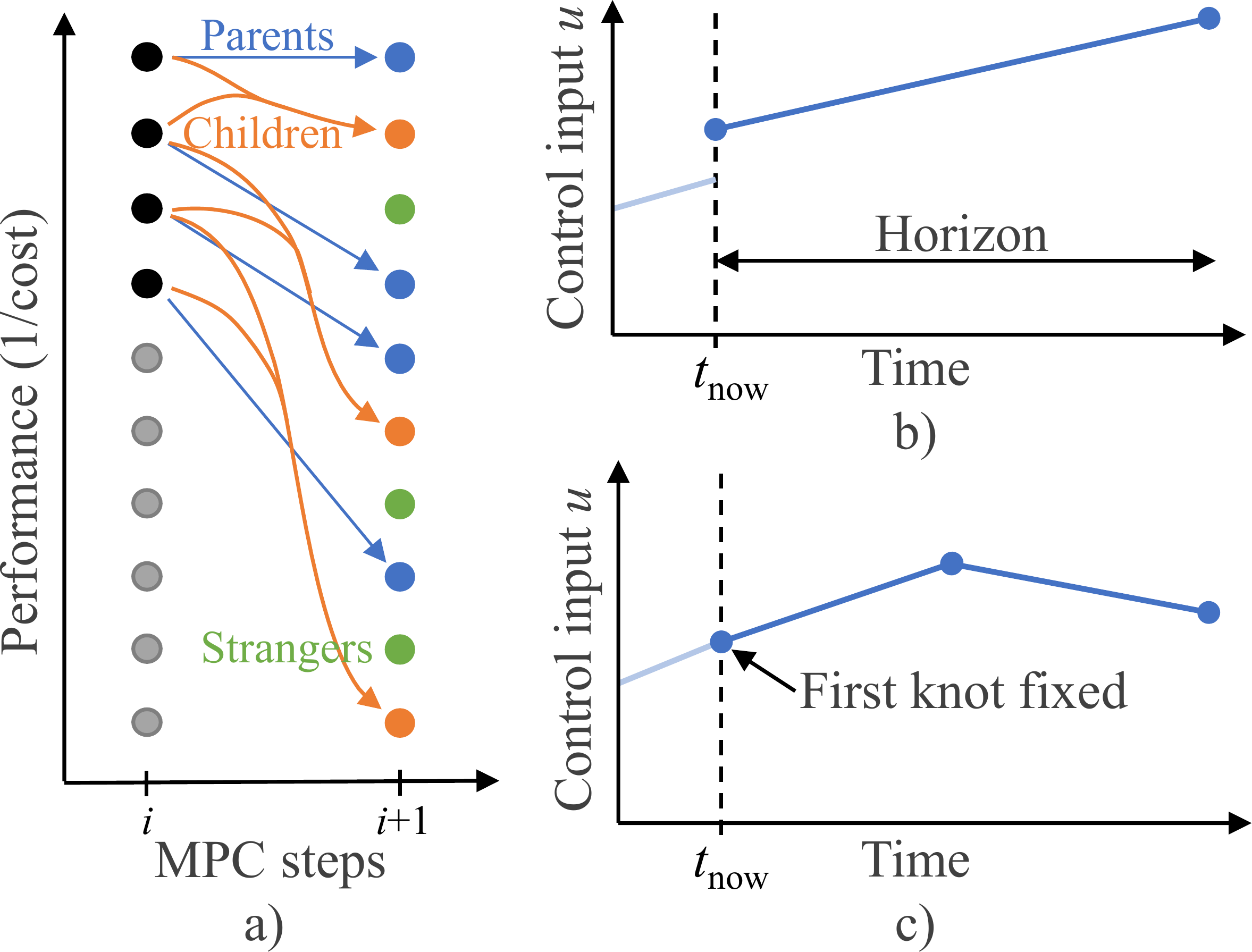}
\caption{Visualization of NEMPC algorithm where the dots represent input trajectory candidates (a) and the input parameterization used to reduce the dimension of the solution space with one (b) and two linear segments (c).}
\label{fig_NEMPC_explainer}
\end{figure}

The Nonlinear Evolutionary Model Predictive Control (NEMPC) algorithm, introduced by Hyatt et al.~\cite{hyatt_realtime_2020}, enables the application of MPC to complex dynamical systems, such as robots with a high number of DoFs. NEMPC achieves this by formulating the control problem as an evolutionary algorithm and exploiting parallel computation on a \acs{gpu} for acceleration.
For each control cycle, a population of particles, each representing a control input trajectory candidate, is evaluated over the prediction horizon. The system's future behavior is predicted for each particle, and the associated cost is computed. The particle yielding the lowest cost is applied to the system, and an evolutionary step is applied to generate the population for the next control cycle (see \figref{fig_NEMPC_explainer}a).
In each evolutionary step, the top-performing $n_\ind{parents}$ particles are carried over to the next generation as parents. The remaining population is filled with stranger particles, randomly sampled from a uniform distribution over the input range, and child particles generated through crossover and mutation. During crossover, each input channel of a child inherits its trajectory from one of two randomly selected parents with a probability of \SI{50}{\%}. This is followed by mutation, where Gaussian noise is added to introduce variability.
To reduce the dimensionality of the optimization problem, NEMPC uses a piecewise linear parameterization of input trajectories (see \figref{fig_NEMPC_explainer}b,c). This simplification is essential, as only a limited number of trajectories can be evaluated per timestep. Nevertheless, the NEMPC has demonstrated the ability to generate complex control behaviors~\cite{hyatt_realtime_2020}.
The algorithm, made publicly available in~\cite{cheney_moldy_2024}, is particularly well-suited for use with \ac{nn} models due to their fast, batch-parallel evaluation on the \acs{gpu}. However, a key limitation of this approach is the absence of guarantees regarding optimality or deterministic convergence.

\subsubsection{NEMPC implementation and tuning} \label{sec:NEMPC_tuning}

In this work, the PyTorch NEMPC implementation from \cite{cheney_moldy_2024} was largely adopted, including the piecewise linear parameterization of control inputs introduced in \cite{hyatt_realtime_2020}. However, we introduced two key modifications: First, the sampling of stranger particles was changed from a uniform to a normal distribution centered around the previous control input, with a standard deviation of $\sigma = 0.3 \cdot (u_{\text{max}} - u_{\text{min}})$. This refinement effectively reduces the solution space to a set of more likely solutions, thereby increasing the likelihood that one of the 1000 evaluated trajectories is the optimal one. Second, an adaptive mutation noise scaling was implemented, which adjusts based on the current control error to achieve an adaptive balance of exploration and convergence. 
The optimal-control problem can be rendered as
\begin{subequations}\label{eq_ocp}
    \vspace{-10pt}
    \begin{multline}
          \min_{\bs{u}_0,...,\bs{u}_{N}} \sum_{k=0}^{N} \lVert \bs{x}_\ind{goal}^{(k)}-\hat{\bs{x}}^{(k)} \rVert_{\bs{Q}}^{2} +  \sum_{k=1}^{N-1} \lVert \bs{u}^{(k)}-\bs{u}^{(k-1)} \rVert_{\bs{R}}^{2} + \\ \lVert \bs{u}^{(0)}-\bs{u}_\ind{prev} \rVert_{\bs{R}}^{2}     
    \end{multline} \vspace{-5pt}
\begin{alignat}{2}
 \text{s.\,t.}\ \hat{\bs{x}}^{(0)} &= \bs{x}_{0} \\
   \hat{\bs{x}}^{(k+1)} &= \Tilde{\bs{f}}_\ind{PINN}(t_s,\hat{\bs{x}}^{(k)},\bs{u}_\ind{filt}^{(k)},{c}_\ind{b}), \nquad k=0,...,N-1 \\
  \bs{u}_\ind{min} &\leq \bs{u}_\ind{filt}^{(k)}\leq \bs{u}_\ind{max} , \nquad k=0,...,N-1. \label{eq_ocp_input_constr}
\end{alignat}
\end{subequations}
The pressures $\bs{u}_\ind{filt}$ are calculated from the sampled pressure trajectories $\bs{u}$ using the identified pressure controller model
\begin{equation}
    \bs{u}_\ind{filt} = \bs{g}_\ind{p}(\bs{u},\bs{u}_\ind{meas})
\end{equation}
further described in \sect{sec_param_ident}. The current measured pressure $\bs{u}_\ind{meas}$ is used as the initial value for filtering.

\begin{table}[!b]
    \centering
    \caption{MPC parameters tuned in simulation by Bayesian optimization. The parameters in brackets are used at the test bench.}
    \addtolength{\tabcolsep}{-0.2em}
    \renewcommand{\arraystretch}{1.25}
    \newcolumntype{Y}{>{\centering\arraybackslash}X}
    \newcolumntype{W}[1]{>{\hsize=#1\hsize\centering\arraybackslash}X}
    \begin{tabularx}{\linewidth}{ p{1.4cm} | W{1.05} W{0.6} W{1.2} W{1.15} } 
    \hline
        \multirow{4}{*}{\shortstack[c]{NEMPC\\parameters}}  & {parent quota} & \multicolumn{2}{c}{stranger quota} & {mutation prob}  \\ \cline{2-5}
        & {0.39 (0.3)} & \multicolumn{2}{c}{0.101 (0.3)} & {0.145 (0.4)}  \\ \cline{2-5}
        & \hlchangesthree{$N$} & {$N_\ind{segments}$} & {MNF} & {$\bs{R}$ factor}  \\ \cline{2-5}
        & \hlchangesthree{{5 (8)}} & {1 (1)} & {0.0366 (0.003)} & {2.39 (40)}  \\ \specialrule{.14em}{.07em}{.07em} 
    \end{tabularx}
    \begin{tabularx}{\linewidth}{ p{1.4cm} | Y Y Y Y Y}
        \multirow{2}{*}{$\bs{Q}$ weights} & $q_\ind{velo}$ & $q_\ind{angvel}$ & $q_\ind{force}$ & $q_\ind{forceZ}$ & $q_\ind{torque}$ \\ \cline{2-6}
        & 0.192 & 0.0504 & 0.746 & 14.6 & 0.186  \\ \cline{1-6}
    \end{tabularx}
    \label{tab:mpc_param}
\end{table}
The parameters for the \ac{nempc} algorithm, along with the weighting matrices $\boldsymbol{Q}$ and $\boldsymbol{R}$, were optimized through Bayesian optimization in simulation. 
Specifically, the entire control loop was simulated in Matlab, using the first-principles Cosserat rod model to simulate the real-world \ac{scr} and calling the Python implementation of the \ac{nempc} from Matlab.
In the first optimization step, the \ac{nempc} algorithm parameters were optimized on a setpoint control and a trajectory tracking task using the \ac{rmse} of the normalized states as cost. The tuned parameters are the quota of parents and strangers, the mutation probability, the number of segments, the mutation-noise scaling factor (MNF), the scaling factor of the $\bs{R}$ matrix, and whether the first knot of the input trajectories is fixed to the previous control input. 
By normalizing the states, it was ensured that each state would contribute similarly. 
The prediction horizon was set to $N$ and the number of particles to 1000, as this was the maximum horizon achievable while maintaining real-time capability. The best-performing configuration was found to be one segment, with a non-fixed first knot, as shown in \figref{fig_NEMPC_explainer}b. The remaining parameters are listed in Table~\ref{tab:mpc_param}. Note that the \ac{nempc} parameters needed to be adapted heuristically to achieve the real-world control results in \sect{sec_mpc_exp_real}. For example, the mutation noise factor needs to be lowered to counteract the significantly higher errors due to modelling inaccuracies. The adapted values are added in brackets.

In the second optimization step, the state weights $Q_{i,i}$ were optimized using the same combined setpoint control and trajectory-tracking task and the tuned \ac{nempc} parameters. To reduce the number of parameters, the weights for internal force, moment, velocity, and angular velocity were applied across all three spatial directions and seven nodes, with the exception of forces in the z direction (see Table~\ref{tab:mpc_param}).

\subsubsection{Exp. 4: NEMPC in Simulation} \label{sec_mpc_exp_sim}
\begin{figure}[!tb]
\centering
\includegraphics[width=\linewidth]{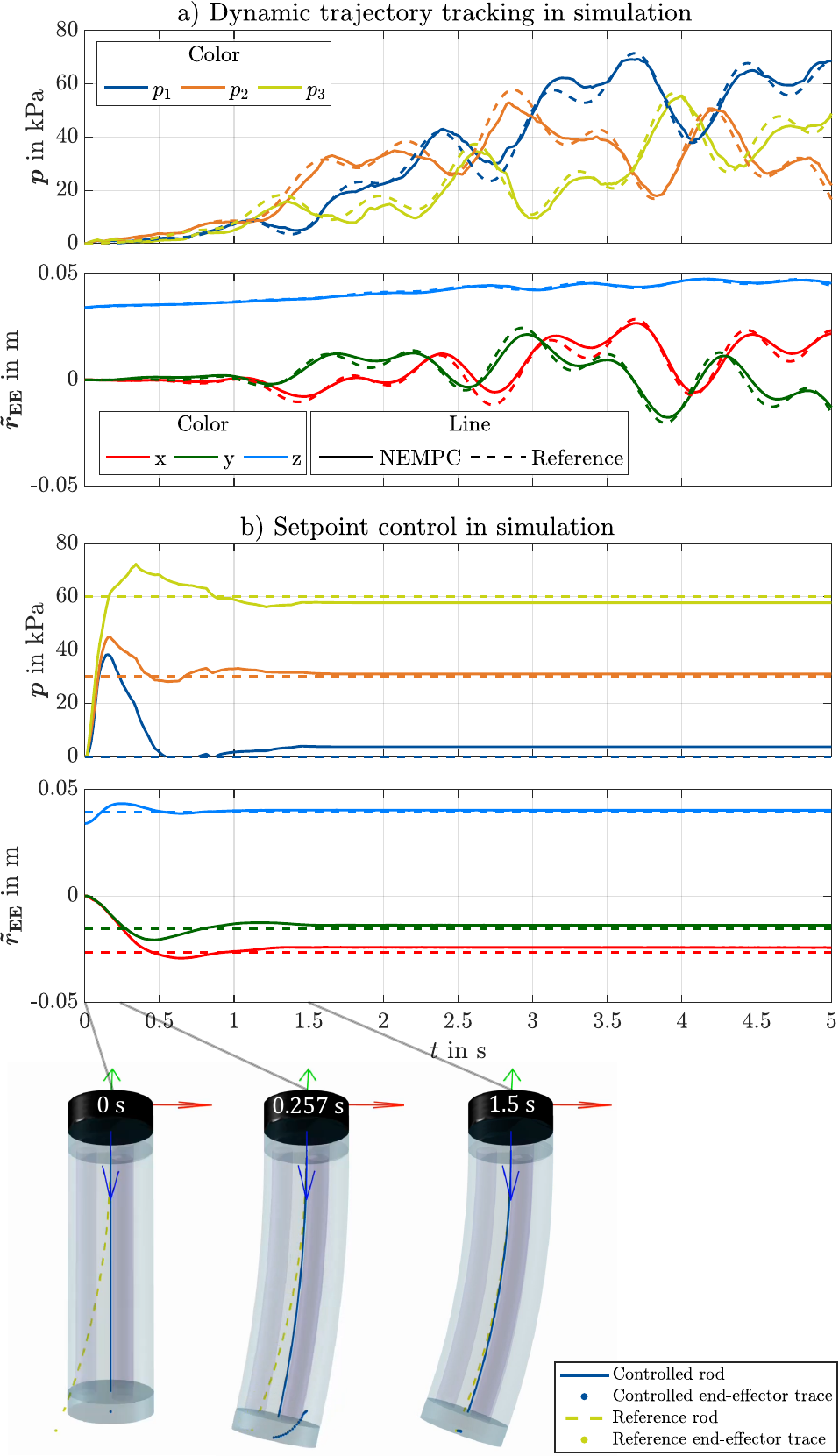}
\caption{Exp. 4: Control results for a) dynamic trajectory tracking and b) setpoint control in simulation \hlchangestwo{with 3D shape plots for illustration.} 
}
\label{fig_MPC_sim_press}
\end{figure}

To investigate the control performance in simulation, the soft actuator's behavior is simulated using the identified Cosserat rod and pressure models. Gaussian noise is added to the end-effector position used for state estimation to emulate realistic conditions. For assessing trajectory-tracking performance, a highly dynamic reference pressure trajectory with oscillations up to \SI{1.5}{Hz} is used (see \figref{fig_MPC_sim_press}a). The corresponding states, calculated via the physics model, serve as the goal trajectory $\bs{x}_\ind{goal}(t)$. 

Although the full Cosserat state\hlchangesthree{, from which the full discrete shape can be reconstructed,} is controlled, we focus primarily on the end-effector position to evaluate the controller's performance, because the high number of states makes visual comparison difficult, and most of the states are not directly measurable.
\hlchangesthree{Nevertheless, we evaluate the shape error using the discrete Fr\'echet distance~\cite{frechet_quelques_1906,danziger_discrete_2026} between the reference and the resulting shape, each consisting of the rod nodes and the end-effector position. 
}
Comparing the end-effector position derived from the desired states and the one resulting from \ac{nempc} control yields an average tracking error (\ac{aed}) of \SI{2.68}{mm}, which is equivalent to \SI{2.1}{\percent} of the actuator's length.
\hlchangesthree{The Fr\'echet distance equals the end-effector error at every time step, i.e., the largest shape deviation occurs at the tip. Hence, the \ac{aed} also represents the mean shape error.}

In addition, setpoint control is tested in simulation with a target state far from the initial configuration (see \figref{fig_MPC_sim_press}b). The system reaches the final position with an error of \SI{2.82}{mm} (\SI{2.2}{\percent} of the actuator's length)\hlchangesthree{, which equals the Fr\'echet distance,} after approximately \SI{1.5}{s}.

This demonstrates the capability of the physics-informed neural model-predictive controller to accurately track trajectories with frequencies of up to \SI{1.5}{Hz} in simulation. The tracking errors of both the full state and the end-effector position remain low. This implies the ability of the MPC to consider the delay of the pressure input and to control complex input trajectories. Moreover, using the same \ac{nempc} parameters, setpoints across the workspace can be reached quickly and with minimal position errors in simulation, confirming the accuracy and robustness of the \ac{pinn}.

\subsubsection{Exp. 5: NEMPC at the Test Bench} \label{sec_mpc_exp_real}
\begin{figure}[!tb]
\centering
\includegraphics[width=\linewidth]{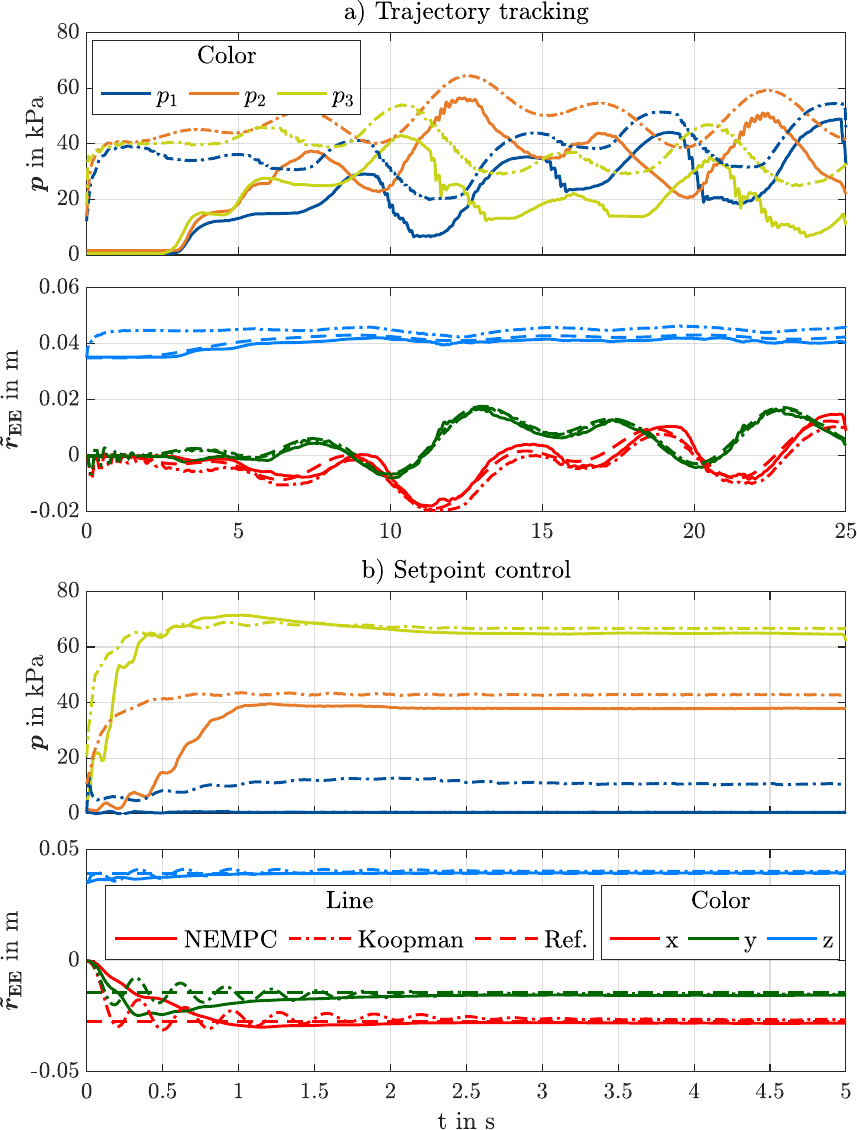}
\caption{Exp. 5: Control result for a) trajectory tracking and b) setpoint control tasks in the real world. The pressures $\bs{p}$ and end-effector position $\Tilde{\bs{r}}_\ind{EE}$ are plotted over time $t$.}
\label{fig_MPC_real}
\end{figure}
This experiment assesses the \ac{nempc}'s performance in the real world using a complex trajectory (Exp.~5a) and a setpoint control task (Exp.~5b). 
\hlchanges{Here, the \ac{nempc} and UKF are used to control the soft actuator in real-time. The experimental setup is described in \sect{sec:SCR+setup}. To ensure that the cumulative computation time for the UKF and MPC is below the sampling time, they are calculated in parallel. This introduces a short delay, as the state estimate is based on measurements from the previous time step, which is considered in the MPC as an input delay of one time step.} \hlchangestwo{The end-to-end computation times of the PINN-based UKF and MPC (calculated in parallel) as well as the physics-model-based UKF and MPC (calculated in parallel) are shown in \tabref{tab:comp_times}. These were tested on the real-time PC (\sect{sec:SCR+setup}) and imply an end-to-end speed-up of \num{12764}.}

\hlchanges{For a comparison to the state of the art, we implemented the dynamic Koopman LQR controller presented by Haggerty et al., which we trained on \SI{15}{min} random step inputs \hlchangesthree{with an equivalent sample frequency of \SI{70}{Hz}} as described in~\cite{haggerty_control_2023}. \hlchangestwo{We selected a pressure range of $[0,90]\,\text{kPa}$ to cover the entire workspace and a hold time of \hlchangesthree{\SI{4}{s}} to allow the oscillations to fully decay.} The controller weights are tuned in simulation, similar to the NEMPC parameters (\sect{sec:NEMPC_tuning}). In contrast to the proposed controller, this one only controls the tip position.}

The reference state trajectory for Exp.~5a is calculated offline using the physics model from a given pressure trajectory with pressure frequencies up to \SI{0.2}{Hz} (see \figref{fig_MPC_real}a). The position visualized as the reference is calculated from the desired states using the physics output model. For the pressure, no reference is plotted because a different pressure trajectory is required to achieve the desired state trajectory in the real world, due to model errors. 
\hlchangesthree{Since no ground truth of the full shape is available at the test bench, the Fr\'echet distance is calculated between the reference shape and the shape estimated by the UKF from the measured tip position, which is the feedback of the \ac{nempc}. Its mean is \SI{2.38}{mm}. }
The achieved tip position \ac{aed} is \SI{2.57}{mm} (x:\,\SI{1.63}{mm}, y:\,\SI{1.22}{mm}, z:\,\SI{1.16}{mm}), which is equivalent to \SI{2.0}{\percent} of the real actuator's length. 
\hlchangesthree{Here, the shape error is smaller than the tip position error because the estimated tip position differs from the measured one.}
\hlchanges{The Koopman controller can control the x and y position with similar errors \hlchangesthree{(x:\,\SI{2.31}{mm}, y:\,\SI{0.85}{mm})}
, but with a considerably larger error in the z direction \hlchangesthree{(\SI{4.29}{mm})} 
, resulting in an \ac{aed} of \hlchangesthree{\SI{5.33}{mm}}.} 

For Exp.~5b, the static state setpoint is calculated using the physics model for a given set of pressures $\bs{p}_\ind{ref} = \trans{[0, 30, 60]}\,\text{kPa}$. The pressure $p_3$ rises quicker than $p_2$, so the target position is reached first in the y- and then in the x-direction (see \figref{fig_MPC_real}b). After about \SI{2}{s}, a steady state with an end-effector position error (\ac{aed}) of \SI{1.69}{mm} (\SI{1.3}{\percent} of the actuator's length) is reached. 
\hlchangesthree{The mean Fr\'echet distance between the reference and the estimated shape from \SI{2}{s} on is \SI{1.57}{mm}.}
\hlchanges{In comparison, the Koopman controller leads to a faster initial setpoint approach. However, it causes a high pressure offset for $p_1$ and a slowly decaying oscillation, resulting in an \ac{aed} of \SI{2.00}{mm} from \SI{2}{s} on.}

These results of Exp.~5a and 5b show that the proposed method can solve the trajectory tracking and setpoint control tasks not only in simulation but also on a real actuator without significant loss in accuracy. \hlchanges{This accuracy surpasses the accuracy of the state-of-the-art Koopman controller.}

\begin{figure}[!tb]
\centering
\includegraphics[width=\linewidth]{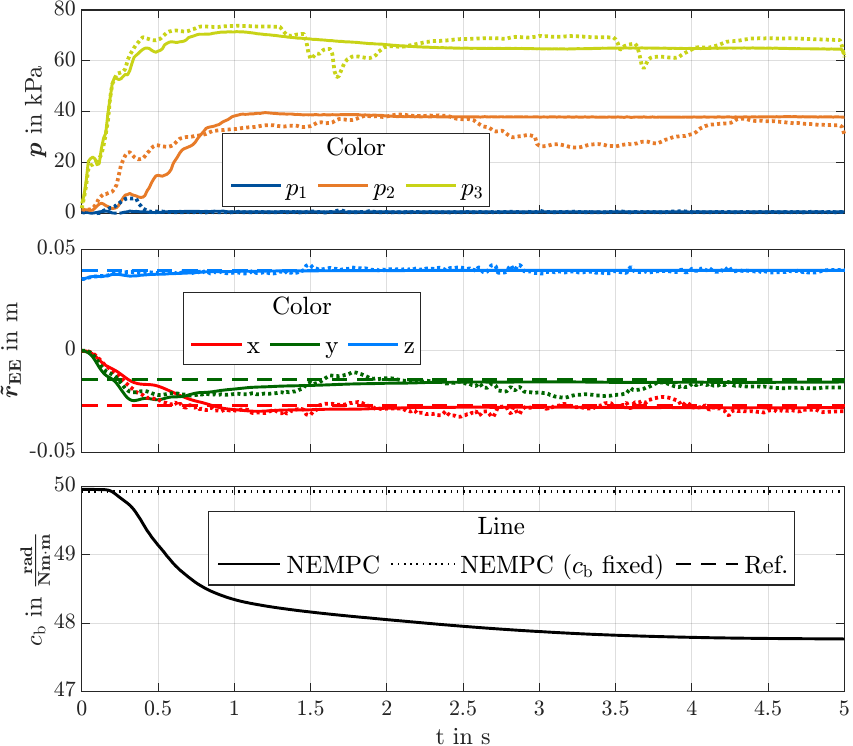}
\caption{Exp. 6: Control result for the NEMPC with and without parameter adaptation for setpoint control in the real world. The pressures $\bs{p}$, end-effector position $\Tilde{\bs{r}}_\ind{EE}$ and bending compliance $c_\ind{b}$ are plotted over time.}
\label{fig_MPC_adapt}
\end{figure}

\hlchanges{
\subsubsection{Exp. 6: NEMPC Adaptivity} \label{sec:Exp6}
Finally, the influence of the parameter adaptivity on control performance is investigated. Therefore, the setpoint control task is performed by the proposed adaptive NEMPC and the same controller with fixed bending compliance $c_\ind{b} = c_\ind{b,ident}=49.9$\,\SI{}{\frac{rad}{Nm\cdot m}} (see \figref{fig_MPC_adapt}). As reported in Exp. 5b, the adaptive controller reaches a steady state after about \SI{2}{s} with a position error of \SI{1.69}{mm}, while the estimated bending compliance reaches a steady state of \SI{47.5}{\frac{rad}{Nm\cdot m}} after \SI{4}{s}. The non-adaptive controller exhibits similar behavior up to \SI{1.5}{s}, but then changes the pressures, presumably trying to reach the setpoint more accurately. This, however, leads to a higher error of \SI{5.10}{mm} from \SI{2}{s} on.}

\begin{figure*}[!htb]
    \centering
    \includegraphics[width=\linewidth]{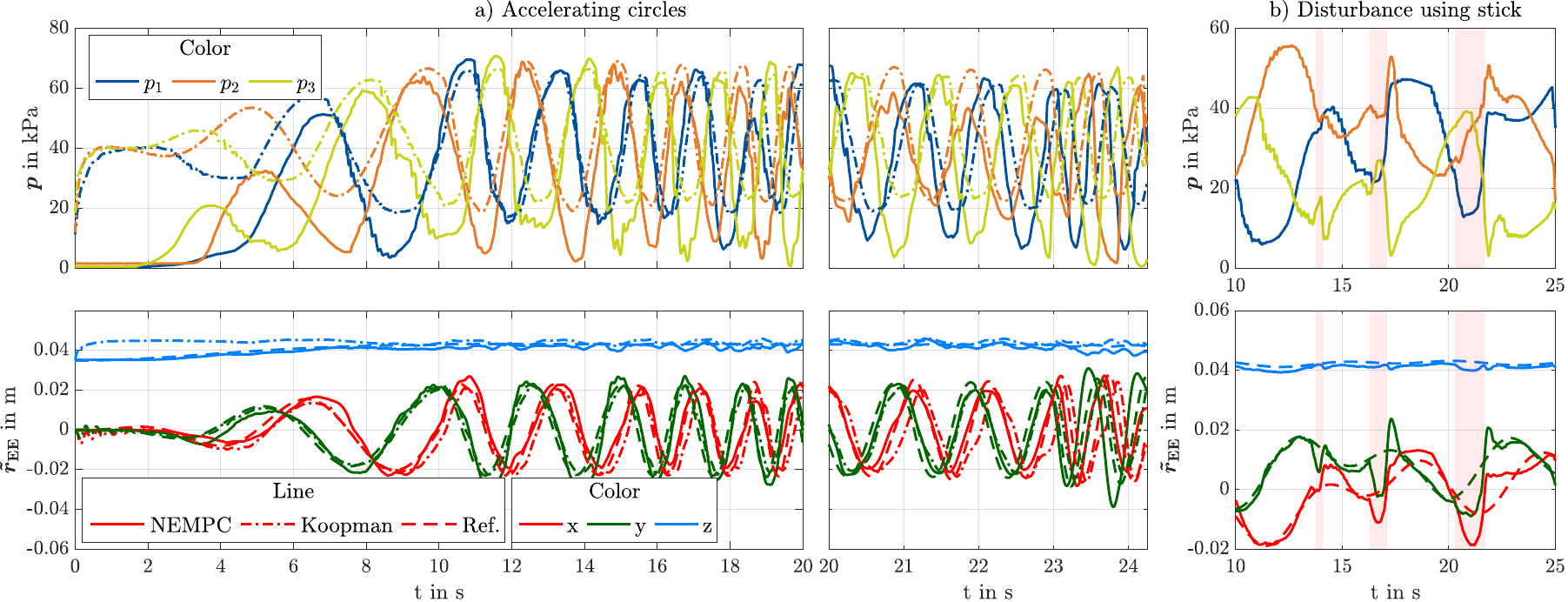}
    \caption{Exp. 7a): Real-world control results for tracking circular motions with increasing frequencies of \{0.2, 0.4, 0.6, 0.8, 1.0, 1.5\}\,Hz \hlchangestwo{and Exp. 7b): Trajectory tracking with disturbances introduced by poking with a stick (marked by the red regions). Exp. 7c) with disturbances caused by a dangling weight attached to the tip is only shown in the supplementary video.}}
    \label{fig:accel_circles}
\end{figure*}
\hlchangestwo{
\subsubsection{Exp. 7: NEMPC Limitations}
To assess the limitations of the proposed controller regarding trajectory speed, a circular trajectory with increasing speeds is tracked in Exp. 7a (see \figref{fig:accel_circles}a) and to investigate the behavior under disturbances, the soft robot is poked with a stick in Exp. 7b. 
In Exp. 7a, the Koopman control results are included for comparison to the state of the art. The frequency is increased every two periods in the following steps: \{0.2, 0.4, 0.6, 0.8, 1.0, 1.5\}\,Hz. Apart from some overshooting, which occurs at \SI{0.4}{Hz} in the x direction and for higher frequencies in the y direction, the trajectory is followed closely. The performance only deteriorates drastically at a trajectory frequency of \SI{1.5}{Hz}. The average end-effector position errors for the given frequencies are \{3.7, 8.7, 8.9, 8.4, 7.3, 18.2\}\,mm. 
\hlchanges{With position errors of \hlchangesthree{\{6.4, 7.9, 10.9, 13.2, 15.3, 21.4\}\,mm}
, the Koopman controller achieves \hlchangesthree{lower accuracy for higher frequencies} and again shows large z errors in the beginning.}}

\hlchangestwo{
\figref{fig:accel_circles}b shows how the soft robot is deflected from its reference trajectory by poking it with a stick in Exp. 7b. The disturbance severity (highlighted in red) increases each time, reaching a maximum of \SI{17.6}{mm}. The controller seems to counteract it by increasing/decreasing the pressures. However, it shows no sign of instability. When the disturbance is gone, it quickly returns to the reference trajectory. \hlchangesthree{During the shown time-frame an average tip position error of \SI{5.2}{mm} is achieved.}

In Exp. 7c, which is included in the supplementary video%
\footnote{\url{https://youtu.be/gNmGwUMiWAc}}, the controller demonstrates \hlchangesthree{empirical} stability even when altering the dynamics by attaching a dangling weight to the robot's tip. \hlchangesthree{The average tip position error measures \SI{2.8}{mm}.}
}


\section{Discussion} \label{sec_discussion}
\hlchangestwo{We discuss our findings with respect to our stated contributions (C1--C4) in connection with our experimental evidence.}

\hlchangestwo{\ref{C1} and \ref{C2}}:
The training of the \ac{ddpinn} as an accurate surrogate model for the Cosserat rod dynamics model is a major contribution to \ac{scr} modeling and control. 
By accelerating the simulation \SI{44800}{} times (batchwise evaluation) while showing almost equal accuracy and no requirement for additional experimental data, it enables real-time control and estimation of the full continuum shape at a frequency of \SI{70}{Hz}.
\hlchangestwo{Both, the quality of the model, as well as the quality of the \ac{ddpinn} are shown by Exp. 2 (\sect{sec_Exp2}).} 

{\hlchangestwo{\ref{C3}}}:
The adaptiveness of the \ac{ddpinn} regarding the bending compliance, furthermore, enables online parameter estimation using the \ac{ukf} as demonstrated in Exp.~3 (\sect{sec:Exp3}). 
This is not possible with previous hybrid or data-driven methods for \ac{scr} control \hlchanges{and can significantly improve the performance, as shown in Exp.~6 (\sect{sec:Exp6})}.

{\hlchangestwo{\ref{C4}}}:
The nonlinear evolutionary model predictive controller based on the \ac{ddpinn} is able to track complex trajectories with an average tip position error below \hlchanges{
\SI{2.1}{\percent} of the actuator length} and quickly realize desired setpoints with a steady-state position error below \hlchanges{\SI{2.2}{\percent} of the actuator length} in simulation \hlchangestwo{(Exp.~4, \sect{sec_mpc_exp_sim})} and real-world experiments \hlchangestwo{(Exp.~4, \sect{sec_mpc_exp_real})}. 
\hlchanges{In a direct comparison of the real-world control performance, we demonstrated that this accuracy surpasses that of the state-of-the-art Koopman controller~\cite{haggerty_control_2023}. 
Our approach reduces the position error by \hlchangesthree{\SI{52}{\percent}} 
for tracking a complex trajectory (Exp. 5a) and by \hlchangesthree{\SI{15}{\percent}} 
for setpoint control (Exp. 5b), compared to the Koopman controller. 
In these scenarios, our adaptive PINN-based approach demonstrates its improved generalizability compared to the Koopman approach.}

The control performance remains acceptable for circular trajectories with oscillation frequencies up to \SI{1.0}{Hz} in the real-world, which is faster than the real-world trajectories tracked in many previous hybrid~\cite{liu_physicsinformed_2024} and \hlchanges{physics-model-based} approaches for soft continuum robot control \cite{katzschmann_dynamic_2019,spinelli_unified_2022,hyatt_model_2020}.
The control frequency of \SI{70}{Hz} is still much higher than the oscillation frequency and therefore, presumably not the reason for the performance loss at higher oscillation frequencies. 
Instead, the pressure controller behavior seems to be dependent on configuration and speed, impeding accurate modeling in the NEMPC. 
The good tracking performance for oscillation frequencies up to \SI{1.5}{Hz} in simulation (Exp.~4a) suggests that the linear parameterization of the NEMPC and the \ac{pinn}'s accuracy are not the cause. 
\hlchanges{This is also suggested by a similar accuracy loss of the Koopman controller at higher oscillation frequencies in Exp.~7a, which uses the same pressure controller.} 
Hence, a different pressure controller could presumably increase the achievable speed even further. 
\hlchangestwo{The robustness of the proposed control approach, even under severe disturbances by poking or adding a dangling weight, is shown in Exp. 7b,c.}
\subsection{Limitations}
Although our approach enables real-time dynamic control of the full shape, defined by the model states, using an accurate Cosserat rod model and allows for adaptation to changes in bending compliance, it also entails several limitations.
The \ac{pinn} approximation is not as accurate in some states, which results in a larger error of the base force. \hlchanges{However, this had no significant impact on the control and state estimation performance.} 
Furthermore, it becomes unstable at inputs above \SI{75}{\kilo\pascal}. 
This could potentially be improved by experimenting with different sampling strategies and trajectories to parameterize the sampling ranges of collocation points.
Furthermore, the \ac{ukf} and the \ac{nempc} require extensive parameter tuning, and the \ac{pinn} must be trained, which requires high computational resources. 
\hlchangestwo{However, in further investigations, training could be significantly accelerated by utilizing the GPU. Using $n_\ind{batch} =$ \SI{20000}{} (with an equal number of collocation points), a fivefold learning rate, and an Nvidia H200 GPU, the entire training time for 3000 epochs could be reduced to below \SI{7}{h}. Note that such large batch sizes are only viable because the network is trained on a pure physics loss. When training on additional experimental data, a much smaller batch size is typically required. Gradient clipping with a maximum gradient norm of \num{1.0} was used here to prevent exploding gradients.} In addition, frameworks like JAX or TensorFlow could be used instead of PyTorch, which demonstrated faster \ac{pinn} training in benchmarks \cite{bafghi_comparing_2024}.
\hlchangesthree{The faster GPU training could enable training with more inputs and more nodes/states for multi-segment robots. From 5 to 11 nodes, the training duration until $\mathcal{L}_\ind{v}=0.11$ is reached scales either quadratically or exponentially ($R^2=0.998$ vs. $R^2=0.997$) with the number of nodes. Reaching \SI{17}{h} for $n_n=11$. Another approach would be to employ modular per-segment models or lower-dimensional strain-based parameterizations (e.g., GVS~\cite{renda_dynamics_2024}). To retain the real-time capability at the test bench for multi-segment robots, the GPU could be upgraded.}
Another limitation is the dependency on the availability of the end-effector position, which necessitates the use of an expensive exteroceptive \hlchanges{motion capture system}. This may not be feasible for many applications. 
Finally, compared to task-space control, planning trajectories in state space is more complex, which may introduce challenges in trajectory planning.

\section{Conclusion} \label{sec_conclusion}
This work combines the adaptability and data efficiency of physics-based models with the fast inference capabilities of data-driven models to enable dynamic and accurate real-time control of soft continuum robots. 
This is achieved by training a \acf{ddpinn} exclusively on a novel collocation-based dynamic Cosserat rod model and deploying it within a nonlinear evolutionary model predictive controller (NEMPC) on the GPU. 
We show that the \ac{pinn} approximates the first-principles model, \hlchangestwo{which is identified on \SI{10}{\second} \hlchangesthree{dynamic data and 42 static data points},} with lower errors than the model's deviation from experimental measurements for the tip position, orientation, and base moment. 
\hlchangestwo{To the best of our knowledge, this is the first time a state-space model of this order has been successfully learned by a \ac{pinn}.}

\hlchangestwo{The \ac{ddpinn} enables the real-time application of state estimation and model predictive control.}
The presented UKF utilizes the \ac{pinn} and its variable bending compliance to achieve estimation of the bending-compliance parameter with an accuracy of \SI{1.2}{\percent} in addition to estimation of the Cosserat rod states.
The controller achieves accurate tracking of dynamic trajectories up to \SI{1.5}{Hz} and setpoint control in simulation, both with end-effector position errors below \SI{3}{mm}. 
On the real \ac{scr}, trajectory tracking and setpoint control are also achieved with position errors below \SI{3}{mm}, \hlchanges{reducing the position error by \hlchangesthree{\SI{52}{\percent}} or \hlchangesthree{\SI{15}{\percent}} respectively, compared to the state-of-the-art Koopman controller~\cite{haggerty_control_2023}}.
Finally, the \hlchangestwo{robustness under disturbances and }tracking of circular trajectories with frequencies up to \SI{1}{Hz} is shown in the real world, which is faster than most soft-continuum-robot controllers. This accurate predictive control of dynamic motions can enable applications like efficient locomotion by utilizing the elastic material as a potential energy storage, and accelerate existing applications like machine inspection.

Future work will focus on enhancing the versatility and robustness of the proposed approach. 
To improve applicability across a wider range of use cases, interoceptive sensors such as \acp{imu}, \hlchanges{bending or base-reaction wrench sensors} can be explored for state estimation, reducing reliance on external sensing systems. 
Furthermore, the adaptiveness of the model could be extended to account for varying tip masses, enabling accurate control for different payloads. 
\hlchanges{Accelerated GPU training could enable the extension to multi-section SCRs. For these, the method’s data efficiency will be even more critical, as purely data-driven methods would require prohibitively large datasets.}
Finally, for operation in contact-rich environments and to leverage the full-shape-control capabilities, we envision incorporating external contact-force estimation, \hlchanges{for example, based on bend sensors}. 
The underlying Cosserat rod model provides a strong foundation for such developments.
In addition, the approach could be transferred to other high-\ac{dof} systems, like humanoid robots, given the \ac{ddpinn}'s good approximation of high-\ac{dof} state-space models.

\bibliographystyle{IEEEtran}
\bibliography{my_bib,IEEEabrv}

\vspace{-33pt}
%

\begin{IEEEbiography}[{\includegraphics[width=1in,height=1.25in,clip,keepaspectratio]{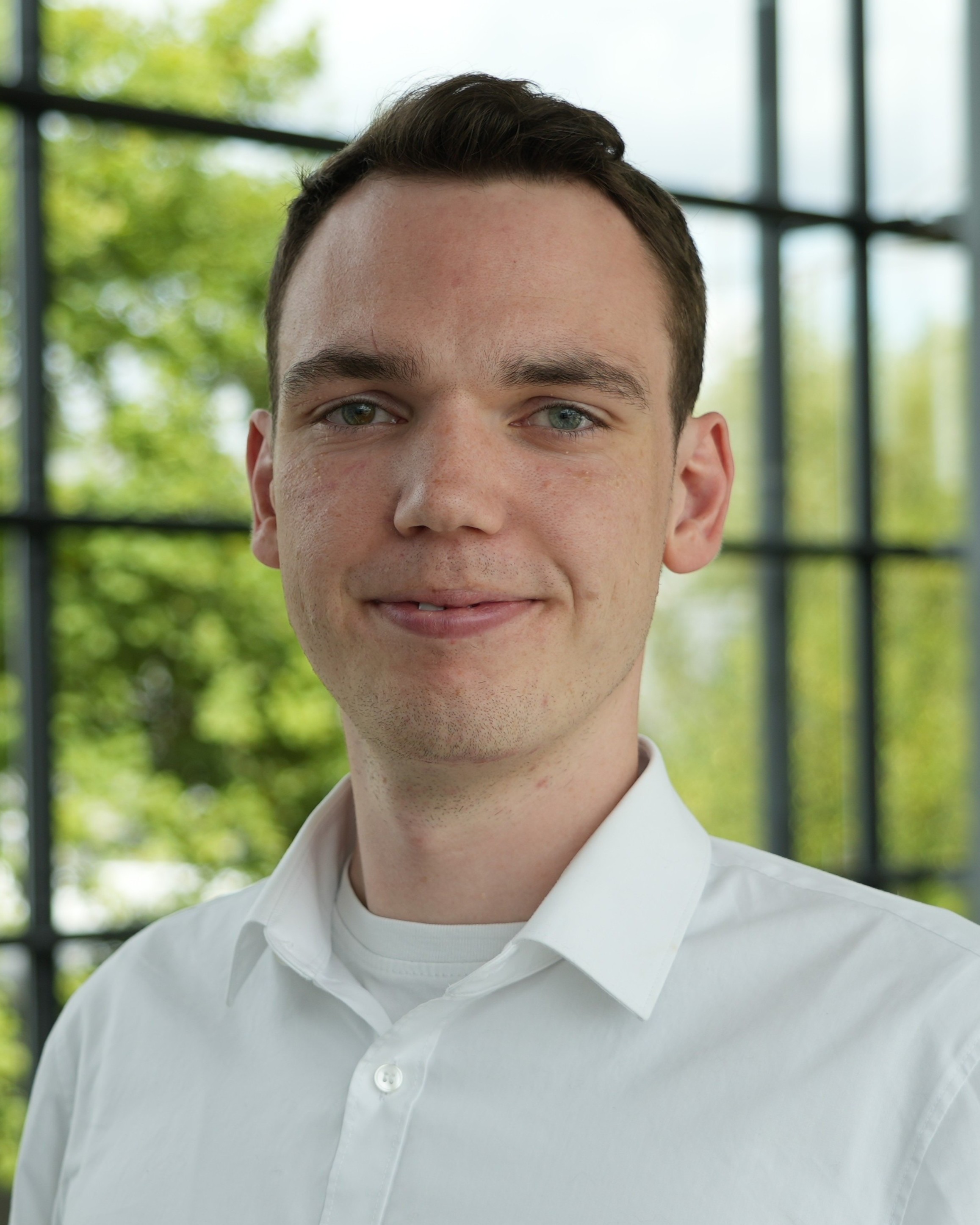}}]{Johann Licher}
(Graduate Student Member, IEEE) received the B.Sc.\ degree and M.Sc.\ degree in mechanical engineering as well as an M.Sc.\ degree in mechatronics and robotics from Leibniz University Hannover, Germany, in 2022 and 2025, respectively.
He is currently working toward the Ph.D.\ degree with the Institute of Assembly Technology and Robotics, Leibniz University Hannover, Germany.
His research interests include the design and control of soft robots and actuators as well as physics-informed machine learning.
\end{IEEEbiography}
\vspace{-33pt}
\begin{IEEEbiography}[{\includegraphics[width=1in,height=1.25in,clip,keepaspectratio]{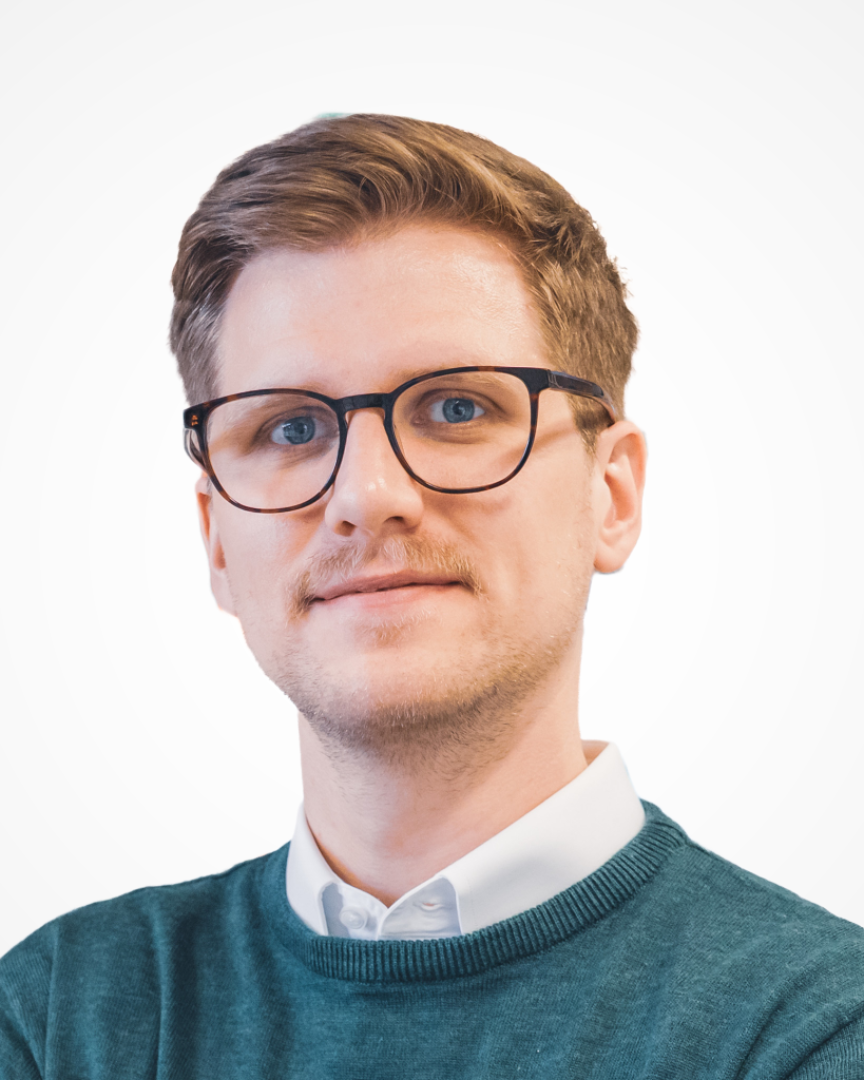}}]{Max Bartholdt}
(Member, IEEE) received the B.Sc.\ and M.Sc.\ in mechatronics engineering, and the Ph.D. degree in mechanical engineering from Leibniz University Hannover, Germany, in 2017, 2019 and 2026 respectively.
He is currently leading a research group in biomedical soft robotics at the Department of Cardiothoracic, Transplantation and Vascular Surgery, Hannover Medical School, Hannover, Germany.
His research interests include modeling, machine learning, state estimation and soft robotics.
\end{IEEEbiography}
\vspace{-33pt}
\begin{IEEEbiography}[{\includegraphics[width=1in,height=1.25in,clip,keepaspectratio]{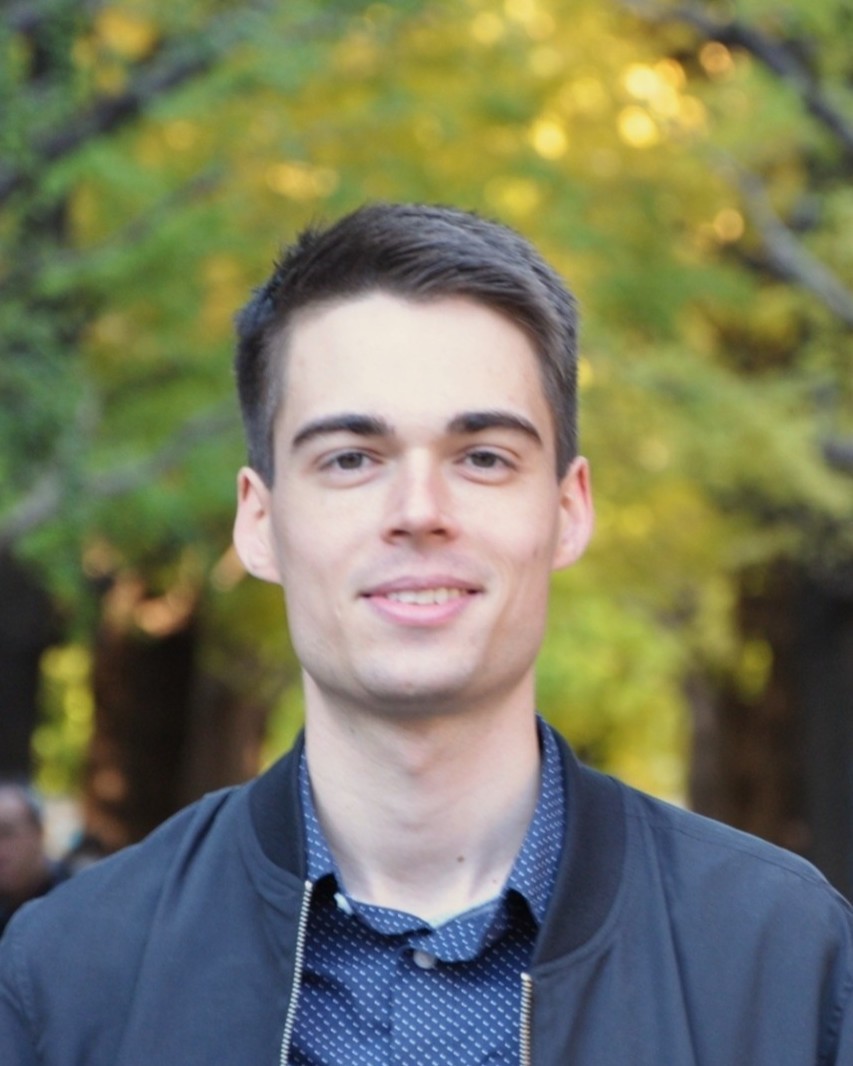}}]{Henrik Krauss}
(Graduate Student Member, IEEE) received the B.Sc. and M.Sc. degrees in mechanical engineering from Leibniz University Hannover, Germany, in 2019 and 2022, respectively, and the M.Sc. degree in engineering from Keio University, Yokohama, Japan, in 2022, within a double-degree program of the T.I.M.E. network. He is currently working toward the Ph.D.\ degree in engineering with the Department of Advanced Interdisciplinary Studies, Graduate School of Engineering, The University of Tokyo, Japan.
His research interests include interpretable machine learning for robot modeling, control, and decision making, with an emphasis on applications to soft robots.
\end{IEEEbiography}
\vspace{-33pt}
\begin{IEEEbiography}[{\includegraphics[width=1in,height=1.25in,clip,keepaspectratio]{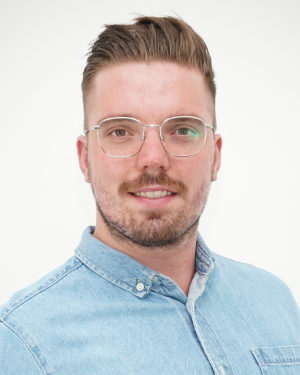}}]{Tim-Lukas Habich}
received the B.Sc., M.Sc.
and the Ph.D. degree in mechanical engineering in 2017,
2020, and 2026 respectively, from Leibniz University Hannover,
Hanover, Germany.
He is currently working as an R\&D engineer at the Carl Zeiss AG.
His research interests include physics-informed
machine learning and learning-based control of
mechatronic systems.
\end{IEEEbiography}
\vspace{-33pt}
\begin{IEEEbiography}[{\includegraphics[width=1in,height=1.25in,clip,keepaspectratio]{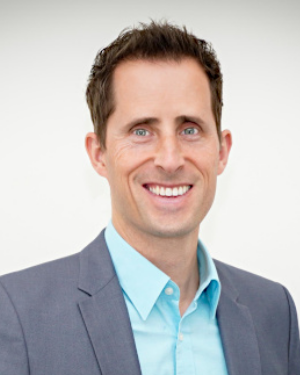}}]{Thomas Seel}
(Member, IEEE) received the Ph.D.
degree in systems and control from Technische Universität
Berlin, Berlin, Germany, in 2016.
He studied engineering cybernetics at Otto von
Guericke University Magdeburg, Germany, and the
University of California, Santa Barbara, CA, USA.
He is currently a Full Professor and Director of the
Institute of Mechatronic Systems, Leibniz University
Hannover, Hanover, Germany, and also member of
the L3S Research Center. His research interests lie
in the development of intelligent and autonomous
mechatronic systems for robotic, automotive, and biomedical applications.
\end{IEEEbiography}
\vspace{-33pt}
\begin{IEEEbiography}[{\includegraphics[width=1in,height=1.25in,clip,keepaspectratio]{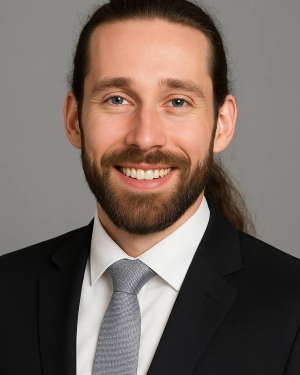}}]{Moritz Schappler}
(Member, IEEE) received the
B.Sc. and M.Sc. degrees in mechatronics engineering
and the Ph.D. degree in mechanical engineering
from Leibniz University Hannover (LUH), Hanover,
Germany, in 2012, 2014, and 2025, respectively.
He is currently leading a research group on Robotic
Systems, Institute of Mechatronic Systems, LUH. His
research interests include design, modeling and control
of robotic and mechatronic systems with a focus
on parallel robots, snake-like robots and methods for
design optimization.
\end{IEEEbiography}



\vfill

\end{document}